\documentclass[preprint,12pt]{elsarticle}

\usepackage{amssymb}
\usepackage{listings}
\usepackage{listings}
\usepackage{algorithm}
\usepackage{algorithmic}
\usepackage{xcolor}
\usepackage{booktabs}
\usepackage{amsmath}
\usepackage{hhline}
\usepackage{subcaption}
\usepackage{multirow}
\usepackage{colortbl}
\usepackage{pifont}

\usepackage{pgfplots}
\pgfplotsset{compat=1.17}

\newcommand{\cmark}{\ding{51}}%
\newcommand{\xmark}{\ding{55}}%

\journal{Knowledge-Based Systems}

\begin{document}

\begin{frontmatter}



\title{Learning Numerical Action Models from Noisy Input Data}

\author[inst1]{Segura-Muros, José Á.\corref{cor1}}\ead{josesegmur@decsai.ugr.es}
\author[inst1]{Fernández-Olivares, Juan}\ead{faro@decsai.ugr.es}
\author[inst1]{Pérez, Raúl}\ead{fgr@decsai.ugr.es}
\cortext[cor1]{Corresponding author}
\affiliation[inst1]{organization={Universidad de Granada},
            addressline={Campus Universitario de Cartuja, C. Prof. Vicente Callao, 3}, 
            city={Granada},
            postcode={18011}, 
            state={Andalusia},
            country={Spain}
}

\begin{abstract}
This paper presents the PlanMiner-N algorithm, a domain learning technique based on the PlanMiner domain learning algorithm. The algorithm presented here improves the learning capabilities of PlanMiner when using noisy data as input. The PlanMiner algorithm is able to infer arithmetic and logical expressions to learn numerical planning domains from the input data, but it was designed to work under situations of incompleteness making it unreliable when facing noisy input data. In this paper, we propose a series of enhancements to the learning process of PlanMiner to expand its capabilities to learn from noisy data. These methods preprocess the input data by detecting noise and filtering it and study the learned action models learned to find erroneous preconditions/effects in them. The methods proposed in this paper were tested using a set of domains from the International Planning Competition (IPC). The results obtained indicate that PlanMiner-N improves the performance of PlanMiner greatly when facing noisy input data.
\end{abstract}



\end{frontmatter}


\section{Introduction}
In recent years, automated planning field has been greatly improved with very fine solutions. These solutions have been enhanced the planners' performance, its ability to face more complex problems or its flexibility to deal with uncertainty. But, even with these improvements, the idea of differentiating the planning engine from the problem representation is still present. This idea differentiates the planning engine from the problem being approached. Making the planning engines independent from the problems help to design generic planning techniques at the expense of the need for writing the ontology of the given problem before trying to solve it. These ontologies are represented using planning domains that are, usually, handcrafted. The main issue when writing planning domains is that is a lengthy process that requires both, extensive knowledge about the problem to be modelled, as well as a lot of time and effort. Model acquisition techniques have been developed over the years, trying to alleviate this. By defining a series of techniques to learn planning domains automatically from existing processes, model domain learning techniques lessen the requirements to write a planning domain lessen.

Among all the proposals presented in the model acquisition techniques area in the last years, PlanMiner \cite{segura2021planminer}, a domain learner based on various machine learning techniques, presented itself as an algorithm capable of learning the most expressive numerical domains possible from a set of plan traces. For an action of a numerical planning domain, PlanMiner is able to learn STRIPS \cite{fikes_strips:_1971} preconditions and effects enriched with arithmetic and logical PDDL \cite{Ghallab98} expressions. PlanMiner is able to achieve this by defining a methodology that integrates classification \cite{kotsiantis2007supervised} and regression \cite{chatterjee2015regression} techniques to process the input data \textendash formatted as Plan Traces\textendash and generate new information from scratch. With this new information, PlanMiner is capable of inferring the numerical preconditions and effects with a higher level of expressiveness to enrich the output PDDL action models. PlanMiner also was designed to deal with incompleteness and is able to learn planning domains with a certain level of missing predicates in the input plan traces, but its performance drops when some noise is included in the input data. The different procedures implemented in PlanMiner are adapted to work with lack of data, so they take a series of assumptions that make the learning process difficult under noise situations. This issue hinders the implementation of PlanMiner in more complex learning problems.

In the literature, \cite{delaRosa12} very good solutions can be found that, like PlanMiner, are able to learn planning domains under levels of incompleteness. Among these approaches, we can find ARMS \cite{Yang07} and LAMP \cite{Zhuo10} which obtain planning domains reducing the learning problem to a Max-Sat problem. FAMA \cite{AINETO2019104} that implements a hybrid learning-planning process able to learn domain models by feeding the input with new information on the fly. The LOCM \cite{cresswell_mccluskey_west_2013} family of algorithms uses finite state machines to learn domains from a single set of plans. Among these algorithms, NLOCM \cite{gregory2016domain} improves LOCM and allows the learning of fixed action's costs. And LC\_M \cite{05e9cf2344f74f8499d051d99bb22488} enhances the LOCM algorithm to be able to learn using traces of noisy plans. Among the solutions that can learn planning domains from noisy input, we can find the proposal of Rodriges et al. \cite{rodrigues2010incremental} present a STRIPS planning domain learning algorithm that learns a preliminary model from an initial input data. Then, the algorithm tests and updates the models by adding new input data incrementally. If a new input example contradicts a model the algorithm modifies it to accommodate the new data. Mourao et al. \cite{Mourao12} learn effects of planning actions by fitting a collection of models using support vector machines \cite{sadohara2001learning} and combining them in a single model. This model is used to generate the output STRIPS planning domain. Finally, Pasula et al. \cite{pasula2007learning} considers every example of the input data as noisy and fits a probabilistic planning domain for them. This domain quantifies the actions' effects in order to detect noisy behaviour. 

Few of these solutions are capable of learning under noisy situations, and none of them can learn numerical planning domains. Seeking to fill this gap, in this paper the authors propose a series of improvements to the PlanMiner learning algorithm to expand its capabilities to learn planning domains correctly under certain noise levels. These methods add new functionality to some of PlanMiner's components to increase their noise tolerance. This is achieved by including several improvements for pre-processing and post-processing the data in the learning process. These improvements (a) study the input data and filter those anomalous elements detected, and (b) try to find incongruencies between the learned models. (a) The study of the data helps PlanMiner to eliminate noisy values, as well as grouping similar Non-CRISP values using discretisation techniques. (b) Finally, in the last stage of the learning process, PlanMiner learns a set of classification models and infers the PDDL action models from them. This process has been enhanced to post-process the models and detect mismatches between them in order to improve the output models.

The different processes implemented in this paper were tested with a set of planning domains from the International Planning Competition (IPC). These domains were used to obtain a collection of plan traces that were populated with random noise.  Using these plan traces as input our experiments tried to reproduce the original IPC domains, measuring the similarity using various quality metrics from the literature. The results showed a promising improvement in the results of PlanMiner dealing with noise.

Next lines will describe the PlanMiner algorithm (Section 2), the input data needed to make it work and the characteristics of the planning domains learned. In section 3 the methods implemented to improve PlanMiner's performance when dealing with noisy input data will be explained. Section 4 will present the experimental setup used to test these new methods, it's results and a discussion of them. Finally, section 5 will conclude this document presenting the conclusions drawn from the experimentation process and the future work to further improve our proposal.

\lstset{tabsize=2, label=lst:PMIn, caption={Extract from an input plan trace from a Rovers planning problem}, escapeinside={(*}{*)}, basicstyle=\footnotesize, breaklines=true}

\begin{minipage}{\linewidth}
\centering
\begin{lstlisting}[mathescape=true]
#(*\underline{Actions}*)
[(*\textbf{start}*)][(*\textbf{end}*)] ((*\textit{action}*))
[(*\textbf{0}*)][(*\textbf{1}*)] (*\textit{(goto rov1 wp1 wp2)}*) 
[(*\textbf{1}*)][(*\textbf{2}*)] (*\textit{(goto rov1 wp2 wp3)}*) 

#(*\underline{States}*)
[(*\textbf{index}*)] ((*\textit{predicates}*))
[(*\textbf{0}*)] (*\textit{(at rov1 wp1)}*) $\land$ (*\textit{($\lnot$ (at rov1 wp2))}*) $\land$ (*\linebreak*)(*\textit{($\lnot$ (at rov1 wp3))}*) $\land$ (*\textit{($\lnot$ (scanned wp3))}*) $\land$ (*\linebreak*)(*\textit{(= (bat\_usage rov1) 3)}*) $\land$ (*\textit{(= (energy rov1) 450)}*) $\land$ (*\linebreak*)(*\textit{(= (dist wp1 wp2) 50)}*) $\land$ (*\textit{(= (dist wp2 wp3) 80)}*)
[(*\textbf{1}*)] (*\textit{($\lnot$ (at rov1 wp1))}*) $\land$ (*\textit{(at rov1 wp2)}*) $\land$ (*\linebreak*)(*\textit{($\lnot$ (at rov1 wp3))}*) $\land$ (*\textit{($\lnot$ (scanned wp3))}*) $\land$ (*\linebreak*)(*\textit{(= (bat\_usage rov1) 3)}*) $\land$ (*\textit{(= (energy rov1) 300)}*) $\land$ (*\linebreak*)(*\textit{(= (dist wp1 wp2) 50)}*) $\land$ (*\textit{(= (dist wp2 wp3) 80)}*)
[(*\textbf{2}*)] (*\textit{(at rov1 wp3)}*) $\land$ (*\textit{($\lnot$ (at rov1 wp1))}*) $\land$ (*\linebreak*)(*\textit{($\lnot$ (at rov1 wp2))}*) $\land$ (*\textit{($\lnot$ (scanned wp3))}*) $\land$ (*\linebreak*)(*\textit{(= (bat\_usage rov1) 3)}*) $\land$ (*\textit{(= (energy rov1) 60)}*) $\land$ (*\linebreak*)(*\textit{(= (dist wp1 wp2) 50)}*) $\land$ (*\textit{(= (dist wp2 wp3) 80)}*)
\end{lstlisting}
\end{minipage}

\section{The \textit{PlanMiner} algorithm}
The PlanMiner \cite{segura2021planminer} algorithm was designed to learn numerical planning domains under situations of incompleteness. For this purpose, PlanMiner defines a methodology that concatenates a series of Machine Learning techniques to, starting from a set of plan traces (Listing \ref{lst:PMIn}) learn a collection of PDDL action models with relational and arithmetic expressions in its effects and preconditions (Listing \ref{lst:PMOut}).

An overview of PlanMiner's workflow can be observed in Algorithm \ref{alg:PM}. Summarising, PlanMiner's input is transformed into a set of datasets (one for each action to be modelled in the output domain) and applies a collection of techniques to them to discover new information and use it to learn relational and arithmetic expressions. Input plan traces are decomposed in a collection of \textit{state transition \textit{($s_{1}$, a, $s_{2}$)}.}, while $s_{1}$ is a snapshot of the world just before executing the action (a \textit{pre-state}), $\mathit{s_{2}}$ is a observation of the world just after executing the action (a \textit{post-state}), and $\mathit{a}$ and action of a plan trace. The state transitions are grouped by the name of its actions \textit{a}, so, given those actions, their related states are used to create a dataset. These datasets display the pre-states and post-states as their instances, with its predicates as the datasets' attributes. Finally, to each instance of the dataset, PlanMiner assigns a class label according to its relation with the action associated with it (either pre-state or post-state).

\begin{algorithm}[tb]
\caption{PlanMiner algorithm overview}
\label{alg:PM}
\textbf{Input} PT: Set of Plan Traces\\
\textbf{Output} AM: Set of Action Models
\begin{algorithmic}[1] 




\STATE Initializes \textit{stDict} as dictionary
\FORALL {Plan trace \textit{pt} in PTs}
    \FORALL {Different action \textit{act} in the \textit{pt}}
        \STATE Extract state transitions \textit{st} of \textit{act} in \textit{pt}
        \STATE \textit{stDict}[\textit{act}] $\leftarrow$ \textit{stDict}[\textit{act}] $\cup$ \textit{st}
    \ENDFOR
\ENDFOR

\FORALL {key \textit{act} in \textit{stDict}}
    \STATE \textit{dat} $\leftarrow$ dataset created using \textit{stDict}[\textit{act}]
    \STATE Infer new knowledge from \textit{dat} and add it
    \STATE Fit a classification model \textit{cModel} using \textit{dat} as input
    \STATE Generate action model \textit{am} from \textit{cModel}
    \STATE AM $\leftarrow$ AM  $\cup$ \textit{am}
\ENDFOR

\STATE \textbf{return} AM
\end{algorithmic}
\end{algorithm}
PlanMiner takes the datasets and enriches them with new information in order to let the learning process to infer a set of numerical and logical expressions. To achieve this, PlanMiner makes use of symbolic regression techniques \cite{var1998multivariate} as the core of the process. These techniques allow PlanMiner to search in the space of arithmetic expressions for the one that best fits a set of target values. PlanMiner detects which numerical predicates evolve when a given action is executed, and then, using a symbolic regressor tries to fit an arithmetic expression that explains these changes. This arithmetic expression uses the rest of numerical predicates of the dataset as variables and a finite set of integers. PlanMiner implements the symbolic regressor as an algorithm of informed graph search \cite{hart1968formal}. The symbolic regression algorithm codes as states the arithmetic expressions and incrementally constructs new expressions from them. PlanMiner creates relational expressions from the numerical predicates of the dataset by selecting a pair of numerical predicates and studying how they behave. Using some expert knowledge PlanMiner is able to detect when a relation between two predicates may be useful, filtering spurious relations. Finally, once a suitable expression (arithmetic or relational) for an action is found, it is included in the given dataset.

Once the datasets have been enriched with new information they are introduced in a classification \cite{kotsiantis2007supervised} algorithm to extract from them a series of features that models the states assigned to a classification label. These features represent a meta-state for the pre-state and for the post-states. The meta-states contains the minimum information to define every pre-state or post-state of a given action. PlanMiner uses a rule-based classification algorithm named NSLV \cite{gonza2009}. NSLV is based on a steady-state genetic algorithm to build the set of classification rules that bests fits a set of input instances. The structure of the rules learned by NSLV is:
$$\mbox{\textbf{IF} } C_1 \mbox{ and }  C_2 \mbox{ and} \ldots  \mbox{and } C_m \mbox{ \textbf{THEN} } Class \mbox{ is } B $$
$$\mbox{ with \textbf{weight w}}$$

\noindent where $\mathit{C_i}$ is a statement "$\mathit{X_n \mbox{ is } A}$", with $\mathit{A}$ a value of the domain of the attribute $\mathit{X_n}$. Each $\mathit{C_i}$  is a feature of the examples that model the $\mathit{B}$. The values $\mathit{A}$ of each variable depends on the type of attribute: True or False for logical attributes or a real number for numerical attributes. The rules are weighted by counting the percentage of instances of the dataset covered by it. In a noise-free environment, NLSV would output two rules: one to model the pre-states and other to model the post-states. 

From these meta-states, is trivial to PlanMiner the task of code the preconditions and effects of the actions. First, the preconditions of the actions are extracted directly by translating the characteristics of the rule that models the "pre-state" class to the PDDL format. Second, the effects of the actions are calculated by comparing the metamodels of the before and after state and extracting their differences. These differences are taken as the necessary steps to transform the pre-state meta-model into a post-state meta-model, that is, the effects of the action.

\lstset{tabsize=2, label=lst:PMOut, caption={Example of learned action by PlanMiner}, escapeinside={(*}{*)}, basicstyle=\footnotesize, captionpos=b, breaklines=true}
\begin{minipage}{\linewidth}
\begin{lstlisting}[mathescape=true]
Action:
  GOTO (?arg_0 - rover ?arg_1 ?arg_2 - wp)
Precondition:
  (at ?arg_0 ?arg_1)
  ((*\textgreater*)= (energy ?arg_0)
    (* (**)(dist ?arg1 ?arg2) (bat_usage ?arg0))
  )
Effects:
  ((*$\lnot$*) (at ?arg_0 ?arg_1))
  (at ?arg_0 ?arg_2)
  (decrease (energy ?arg_0)
    (* (**)(dist ?arg1 ?arg2) (bat_usage ?arg0))
  )
\end{lstlisting}
\end{minipage}

\section{PlanMiner-N}
PlanMiner-N includes two new processes to the pipeline presented in the previous section. These processes presented extend the PlanMiner algorithm to improve its resilience against noise. Such methods were designed with the philosophy of enriching the learning pipeline without modifying its key parts. These new steps developed are the filtering of the noisy input data and the refinement of the meta-states. In the following lines, we will explain these steps in detail, illustrating the whole process with examples taken from domain \textit{Rovers} (defined in Listing \ref{lst:PTnoisy}). This plan traces is a modified noisy trace from the example trace of Listing \ref{lst:PMIn}. In this example noisy trace, the erroneous elements are highlighted in bold. 

\begin{algorithm}[tb]
\caption{PlanMiner-N Algorithm overview}
\label{alg:PMN}
\textbf{Input}: PT: Set of Plan Traces\\
\textbf{Output}: AM: Set of Action Models
\begin{algorithmic}[1] 



\STATE Initializes \textit{stDict} as an empty dictionary
\FORALL {Plan trace \textit{pt} in PTs}
    \FORALL {Different action \textit{act} in the \textit{pt}}
        \STATE Extract state transitions \textit{st} of \textit{act} in \textit{pt}
        \STATE \textit{stDict}[\textit{act}] $\leftarrow$ \textit{stDict}[\textit{act}] $\cup$ \textit{st}
    \ENDFOR
\ENDFOR

\FORALL {key \textit{act} in \textit{stDict}}
    \STATE \textit{dat} $\leftarrow$ dataset created using \textit{stDict}[\textit{act}]
    \STATE Filter noise from \textit{dat} \textcolor{teal}{\texttt{\#Step added in PlanMiner-N}}
    \STATE Detect new features from \textit{dat} and extend \textit{dat} with them
    \STATE Fit a classification model \textit{cModel} using \textit{dat} as input
    \STATE Refine \textit{cModel} \textcolor{teal}{\texttt{\#Step added in PlanMiner-N}}
    \STATE Generate action model \textit{am} from \textit{cModel}
    \STATE AM $\leftarrow$ AM  $\cup$ \textit{am}
\ENDFOR

\STATE \textbf{return} AM
\end{algorithmic}
\end{algorithm}

\subsection{PlanMiner-N Overview}
Algorithm \ref{alg:PMN} shows PlanMiners-N's general workflow, highlighting the changes introduced to PlanMiner's original workflow. PlanMiner-N modifies PlanMiner original contribution to enable the latter to operate under noisy input data situations. As said earlier, PlanMiner-N implements two new steps in the original pipeline of PlanMiner, performing in each one the following tasks:

\begin{enumerate}
    \item \textbf{Plan traces noise filtering.} Just after storing the information of the plan traces in a dictionary whose keys are the action names and their values a set of associated state transitions, this process is applied. This procedure (step 10 of Algorithm \ref{alg:PMN}) aims to clean the input information. Depending on the type of data found in a dataset being addressed (either predicates or numerical fluents), a different process is applied. A noise filter based on statistical filtering is applied to the predicates, which depends on a frequency threshold that determines whether a predicate is noisy or not. Regarding numerical fluents, we assume that the noise produced is random, so they are discretised and smoothed to reduce fluctuations in their values. This must be done so that the other components of the learning pipeline can perform their work to prevent the influence of the noisy values in its output.
    \item \textbf{Meta-state refinement.} As explained above, a meta-state is a characterisation of the predicates/fluents that can be found in either the pre-state or the post-state of an action. In the case of information without noise, each characterisation is directly obtained and represented as the antecedent of a single rule. However, when addressing noisy datasets, the classification model initially obtained (step 12 of Algorithm \ref{alg:PMN}) may contain several characterisations (i.e. it can found several rules) for either the pre-state or post-state of an action. The Meta-state refinement (step 13 of Algorithm \ref{alg:PMN}) solves this, enforcing the constraint that only two meta-states are needed, by combining the elements of the rules intially learned.
\end{enumerate}

\subsection{Plan traces noise filtering}
The noise filtering process of PlanMiner-N aims to detect and delete anomalous erroneous information from the input data. The original PlanMiner algorithm takes the input data and preprocesses it to adjust the format of the data (i.e. dataset extraction) and to enrich it (i.e. discovery of new knowledge), PlanMiner-N introduces a new element to that preprocessing with the filtering of the input data. This new element is included between the two previous ones, and, as previously mentioned, its purpose is to reduce or alleviate as much as possible the noise problems that the input data may have. 

Due to the nature of the information contained in the input data, we need to apply different techniques to it, since the noise treatment for the nominal values of the predicates is different from the noise treatment of the real values of the fluents. The only type of noise that affects predicates is outliers, while fluents are also affected by random noise. In the example noisy trace of Listing \ref{lst:PTnoisy} we can see an outlier in the logical predicates in predicate $\mathit{(at\ rov1 \ wp2)}$ of state \texttt{[0]}, an example of random noise in the numerical predicates in fluent $\mathit{(=(at\ rov1)\  3.25)}$ of the same state, and an example of outlier in the fluents in element $\mathit{(=(dist\ wp2\ wp3) \ 380)}$ of state \texttt{[2]}. This problem with the different types of noise conditions the techniques defined in PlanMiner-N, causing the input data preprocessing step to be applied differently for each type of input data, as there is no jack-of-all-trades preprocessing technique to deal with noise.

On the one hand, for noise in the logical predicates, PlanMiner-N implements statistical filtering in order to detect outliers and eliminate them. This filtering studies the distribution of the different truth values of each predicate along the traces, counting their frequency of occurrence. Then, if a truth value has an anomalously low frequency of occurrence, it is marked as noisy and removed. This process can be implemented because there are no conditional or stochastic behaviours in the learned actions. If such behaviours existed, we would not be able to discern between those outlier truth values or those that are related to an infrequent, but correct, non-deterministic behaviour. 

\lstset{tabsize=2, label=lst:PTnoisy, caption={Example of a noisy plan trace}, escapeinside={(*}{*)}, basicstyle=\footnotesize, breaklines=true, captionpos=b}
\begin{minipage}{\linewidth}
\centering
\begin{lstlisting}[mathescape=true]
#(*\underline{Actions}*)
[(*\textbf{0}*)][(*\textbf{1}*)] (*\textit{(goto rov1 wp1 wp2)}*) 
[(*\textbf{1}*)][(*\textbf{2}*)] (*\textit{(goto rov1 wp2 wp3)}*) 

#(*\underline{States}*)
[(*\textbf{0}*)] (*\textit{(at rov1 wp1)}*) $\land$ (*\textbf{ \textit{(at rov1 wp2)}}*) $\land$ (*\linebreak*)(*\textit{($\lnot$ (at rov1 wp3))}*) $\land$ (*\textit{($\lnot$ (scanned wp3))}*) $\land$ (*\linebreak*)(*\textbf{\textit{(= (bat\_usage rov1) 3.25)}}*) $\land$ (*\textit{(= (energy rov1) 450)}*) $\land$ (*\linebreak*)(*\textit{(= (dist wp1 wp2) 50)}*) $\land$ (*\textit{(= (dist wp2 wp3) 80)}*)

[(*\textbf{1}*)] (*\textit{($\lnot$ (at rov1 wp1))}*) $\land$ (*\textit{(at rov1 wp2)}*) $\land$ (*\linebreak*)(*\textit{($\lnot$ (at rov1 wp3))}*) $\land$ (*\textbf{\textit{(scanned wp3)}}*) $\land$ (*\linebreak*)(*\textit{(= (bat\_usage rov1) 3)}*) $\land$ (*\textbf{\textit{(= (energy rov1) 299)}}*) $\land$ (*\linebreak*)(*\textit{(= (dist wp1 wp2) 50)}*) $\land$ (*\textit{(= (dist wp2 wp3) 80)}*)

[(*\textbf{2}*)] (*\textbf{\textit{($\lnot$ (at rov1 wp3))}}*) $\land$ (*\textit{($\lnot$ (at rov1 wp1))}*) $\land$ (*\linebreak*)(*\textit{($\lnot$ (at rov1 wp2))}*) $\land$ (*\textit{($\lnot$ (scanned wp3))}*) $\land$ (*\linebreak*)(*\textit{(= (bat\_usage rov1) 3)}*) $\land$ (*\textit{(= (energy rov1) 60)}*) $\land$ (*\linebreak*)(*\textit{(= (dist wp1 wp2) 50)}*) $\land$ (*\textbf{\textit{(= (dist wp2 wp3) -8000)}}*)
\end{lstlisting}
\end{minipage}

On the other hand, for numerical predicates, PlanMiner-N bases the noise filtering on a discretisation technique \cite{garcia2015data} that groups the different elements of a fluent under a series of discrete labels that replace them. In ML, discretisation is the process by which a set of continuous variables is transformed into a finite set of discrete variables. The benefits of discretisation \cite{liu2002discretization} include categorising data for the sake of understandability, reducing the number of possible values of an attribute to improve the performance of ML algorithms, or, most relevant to the solution presented in this section, ``smoothing'' the discretised data. This smoothing process reduces fluctuations in the input data caused by random noise. By gathering similar elements under a single label, we do not only reduce the random noise of the data, but we can also isolate those data that are not similar to any other, i.e., the outliers.

This new preprocessing step is a powerful tool that can improve greatly the performance of the whole learning process, but, its major drawback is that it directly affects the execution of PlanMiner-N, increasing the amount of time that it requires to work. The preprocessing step is applied directly to the datasets extracted from the plan traces. The examples designed to illustrate the new methods start from the dataset of Table \ref{tab:datasetN}, a dataset created from the plan trace presented in Listing \ref{lst:PTnoisy}. In this table, in the \textit{(bat\_usage ?arg1)} attribute, examples of outliers (-4) or random noise (5.05) can be observed.

\begin{table}
\setlength{\aboverulesep}{-1.0pt}
\setlength{\belowrulesep}{0.0pt}
    \begin{center}
    \resizebox{\textwidth}{!}{
    \begin{tabular}{cccccc|c}
        \toprule
        (at ?arg1 ?arg2) & (at ?arg1 ?arg3) & (bat\_usage ?arg1) & (energy ?arg1) & (dist ?arg2 ?arg3) & (scanned ?arg3) & \textit{Class} \\ 
        \midrule
        \textit{True} & \textit{True} & 3.25 & 450 & 50 & \textit{MV} & $pre-state$ \\
        \textit{False} & \textit{True} & 3 & 299 & 50 & \textit{MV} & $post-state$ \\
        \textit{True} & \textit{False} & 3 & 300 & -8000 & \textit{True} & $pre-state$\\
        \textit{False} & \textit{False} & 3 & 6000 & 86 & \textit{False} & $post-state$\\
        \textit{True} & \textit{False} & 3 & 230 & 75 & \textit{MV} & $pre-state$ \\
        \textit{False} & \textit{True} & 3 & 5 & 75 & \textit{MV} & $post-state$ \\
        \textit{True} & \textit{True} & 2.97 & 400 & 35 & \textit{True} & $pre-state$ \\
        \textit{False} & \textit{True} & 3 & 295 & 33 & \textit{False} & $post-state$ \\
        \textit{True} & \textit{False} & 5 & 400 & 75 & \textit{False} & $pre-state$ \\
        \textit{False} & \textit{True} & 5.05 & -50 & 75 & \textit{True} & $post-state$ \\
        \textit{True} & \textit{False} & 5 & 500 & 50 & \textit{False} & $pre-state$ \\
        \textit{True} & \textit{True} & 5 & 250 & 50 & \textit{False} & $post-state$ \\
        \textit{True} & \textit{False} & 3 & 315 & 1005 & \textit{True} & $pre-state$ \\
        \textit{False} & \textit{True} & 3 & -0.5 & 105 & \textit{True} & $post-state$ \\
        \textit{True} & \textit{False} & -4 & 500 & 80 & \textit{MV} & $pre-state$ \\
        \textit{False} & \textit{True} & 5 & 100 & 80 & \textit{MV} & $post-state$ \\
        \textit{True} & \textit{False} & 3 & 46 & 15 & \textit{False} & $pre-state$ \\
        \textit{False} & \textit{True} & 3 & 10001 & 15 & \textit{False} & $post-state$ \\
        \bottomrule
    \end{tabular}
    }
    \end{center}
    \caption{Noisy dataset associated with the \textit{(goto ?arg1 ?arg2 ?arg3)} action.}
    \label{tab:datasetN}
\end{table}

\subsubsection{Logical values noise treatment}
Once the datasets have been created, the first action done by PlanMiner-N is to tackle noise in the logical attributes in the dataset. This is performed by PlanMiner-N by implementing a statistical filter to detect anomalies in the data. This filter is implemented over a collection of frequency tables that survey how the information is distributed in the dataset. These frequency tables measure the importance that each value has in a given attribute. In the case of finding an anomaly in these measures, it is filtered out and erased from the dataset. An anomaly is a value with an abnormally low importance.

\begin{algorithm}[tb]
\caption{Statistical noise filter overview}
\label{alg:Filt}
\textbf{Input} dat: Dataset\\
\textbf{Output} dat: Dataset
\begin{algorithmic}[1] 
\FORALL{Class label \textit{cLabel} in \textit{dat}}
    \STATE Initializes \textit{subDat} as empty Dataset
    \FORALL {Instance \textit{i} in \textit{dat} whose class is \textit{cLabel}}
        \STATE Include \textit{i} in \textit{subDat}
    \ENDFOR
    \FORALL {Instance \textit{i} in \textit{subDat}}
        \FORALL {attribute \textit{attr} in \textit{subDat}}
            \STATE tCount $\to$ count number of True values of \textit{attr}
            \STATE fCount $\to$ count number of False values of \textit{attr}
            \IF{$\frac{tCount}{tCount+fCount} < $ threshold}
                \STATE Erase all True values in \textit{attr} 
            \ENDIF
            \IF{$\frac{fCount}{tCount+fCount} < $ threshold}
                \STATE Erase all False values in \textit{attr} 
            \ENDIF
        \ENDFOR
    \ENDFOR
    \STATE Update \textit{dat} with the information of \textit{subDat} 
\ENDFOR
\STATE \textbf{return} dat
\end{algorithmic}
\end{algorithm}

\begin{figure}
    \resizebox{.32\textwidth}{!}{
        \begin{tikzpicture}
        \begin{axis} [xbar = .05cm,
            bar width = 12pt,
            xmin = 0,
            xmax = 1,
            ytick = data,
            enlarge x limits = {value = .05, upper},
            enlarge y limits = {abs = .8},
            yticklabels={Post-state,Pre-state},
            ,xlabel=Frequency rate,
            ,ylabel=$\mathit{(at\ ?arg1\ ?arg2)}$,
            extra x ticks = 0.1111,
            extra x tick labels={},
            extra x tick style={grid=major,major grid style={thick,draw=green}}
        ]
         
        \addplot coordinates {(0.111,0) (1,1)}; 
        \addplot coordinates {(0.888,0) (0,1)}; 
         
        \legend {True, False};
 
        \end{axis}
        \end{tikzpicture}
    }
    \hfill
    \resizebox{.32\textwidth}{!}{
        \begin{tikzpicture}
        \begin{axis} [xbar = .05cm,
            bar width = 12pt,
            xmin = 0,
            xmax = 1,
            ytick = data,
            enlarge x limits = {value = .05, upper},
            enlarge y limits = {abs = .8},
            yticklabels={Post-state,Pre-state},
            ,xlabel=Frequency rate,
            ,ylabel=$\mathit{(at\ ?arg1\ ?arg2)}$,
            extra x ticks = 0.1111,
            extra x tick labels={},
            extra x tick style={grid=major,major grid style={thick,draw=green}}
        ]
 
        \addplot coordinates {(0.888,0) (0.222,1)}; 
        \addplot coordinates {(0.111,0) (0.777,1)}; 
         
        \legend {True, False};
 
        \end{axis}
        \end{tikzpicture}
    }
    \hfill
    \resizebox{.32\textwidth}{!}{
        \begin{tikzpicture}
        \begin{axis} [xbar = .05cm,
            bar width = 12pt,
            xmin = 0,
            xmax = 1,
            ytick = data,
            enlarge x limits = {value = .05, upper},
            enlarge y limits = {abs = .8},
            yticklabels={Post-state,Pre-state},
            ,xlabel=Frequency rate,
            ,ylabel=$\mathit{(at\ ?arg1\ ?arg2)}$,
            extra x ticks = 0.1111,
            extra x tick labels={},
            extra x tick style={grid=major,major grid style={thick,draw=green}}
        ]
 
        \addplot coordinates {(0.333,0) (0.5,1)}; 
        \addplot coordinates {(0.666,0) (0.5,1)}; 
         
        \legend {True, False};
 
        \end{axis}
        \end{tikzpicture}
    }

    \caption{Frequency rates of the values of each predicate of the dataset presented in Table \ref{tab:datasetN}. Each graph represent a predicate and its values. The green line represents the threshold that determines whether a value is considered noisy.}
    \label{fig:freqs}
\end{figure}
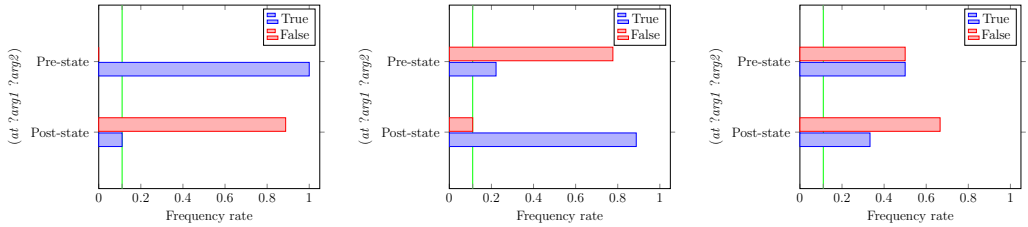

The filter (Algorithm \ref{alg:Filt}) implemented in PlanMiner-N operates as follows: Starting from a single dataset, the filter fixes a class label (either pre-state or post-state) and creates a sub-dataset with only the information of the fixed label. Using this sub-dataset, the filter creates a frequency table for each different logical attribute. These frequency tables measure the number of times a given attribute takes a certain value. For a certain attribute, if the relative number of times a value appears is below a threshold, then it is considered \textit{irrelevant}. Irrelevant attributes are considered as there is no useful information to be extracted from it, thus they are counted as noise. The deletion of a value is realised by selecting its appearances in the attribute's column and replacing them for missing values tokens. Since internally the pipeline components follow the Open World Assumption, the inclusion of a non-determined missing value does not influence the method of operation of the algorithm. Once a sub-dataset has been processed, the other class label is selected and the procedure is repeated. Figure \ref{fig:freqs} presents an example with the frequencies of the predicates from the example dataset presented in Table \ref{tab:datasetN}. In this example, it can be seen graphically the frequency rate of their values. Those values that do not exceed the threshold set by PlanMiner-N (defined by the green bar)  will be eliminated. 

The threshold set to filter noisy elements of the dataset can impact heavily on the performance of the process. In case of dealing with a situation where an attribute displays a non-noisy value with a low appearance rate, a very high threshold value may lead to detecting it as noise, and, therefore, erasing it. The absence of this value may hinder the later learning processes,  including, in fact, an artificial extra noise. On the other side, a low threshold value may work the other way around, setting as ``uncommon examples'', noisy values. This issue would make the filtering process useless.

\subsubsection{Numerical values noise treatment}
After filtering the logical outliers, PlanMiner-N proceeds to process the numerical information in the input data. The variety of data contained in a noisy numerical attribute makes the use of a filter process like the one shown before inviable. 

Even in a noise-free environment, a numerical continuous attribute may display wide range of different values. A statistical filter would not work correctly in this kind of situation, marking as noise the majority of values. PlanMiner-N implements an alternative filtering process (Algorithm \ref{alg:FiltN}) to deal with the noise in the numerical values. This filter takes as input a dataset, selects a numerical attribute, and extracts every element from it. The values extracted are then used as input of a discretisation algorithm that processes and groups them. Finally, the discretised values substitute the original values of the attribute.

\begin{algorithm}[tb]
\caption{Numerical noise filter overview}
\label{alg:FiltN}
\textbf{Input} dat: Dataset\\
\textbf{Output} dat: Dataset
\begin{algorithmic}[1] 
\STATE Initializes \textit{elems} as empty list of numbers
\FORALL {attribute \textit{attr} in \textit{dat}}
    \STATE \textit{elems} $\leftarrow$ \textit{elems} $\cup$ values list of \textit{attr}
\ENDFOR
\STATE\textit{newAttr} $\leftarrow$ discretize \textit{elems}
\STATE Update \textit{dat} with the information of \textit{newAttr} 
\STATE \textbf{return} dat
\end{algorithmic}
\end{algorithm}

The discretisation algorithm must produce a set of finite discrete elements from the collection of values used as input. PlanMiner-N achieves this using a clustering technique. The discretisation process groups the input elements into a set of clusters, and calculates the output discrete set of values as the mean element of each cluster. Given that the characteristics of the data are unknown a priori (the number of data points, their distribution, ...), the algorithm can not use a predefined number of clusters to fit the data. This provokes that PlanMiner-N must find the best number of clusters automatically. PlanMiner-N implements a top-down divisive hierarchical clustering technique \cite{lamrous2006divisive} to achieve this. This type of clustering methodology starts from every value in a single cluster and divides it recursively into smaller clusters. The principle on which these techniques are based is to incrementally create a larger number of clusters that better fit the data. 

\begin{algorithm}[tb]
\caption{Discretization algorithm overview}
\label{alg:HRC}
\textbf{Input} Dat: Set of data points, AC: acceptance criterion\\
\textbf{Output} C: Set of Clusters
\begin{algorithmic}[1] 
\STATE \textit{split} $\leftarrow$ Halve \textit{Dat} using K\_means algorithm
\STATE \textit{C} $\leftarrow$ \{\}
\FORALL{\textit{cluster} in \textit{split}}
\IF {Quality(\textit{cluster}) $>=$ AC}
\IF {$ |\mathit{cluster}| > 1$}
\STATE \textit{C} $\leftarrow$ \textit{C} $\cup$ \textit{cluster}
\ENDIF
\ELSE
\STATE \textit{C} $\leftarrow$ \textit{C} $\cup$ divide \textit{cluster} calling the discretisation process recursively
\ENDIF
\ENDFOR
\STATE \textbf{return} \textit{C}
\end{algorithmic}
\end{algorithm}

The clustering technique defined in PlanMiner-N (see Algorithm \ref{alg:HRC}) follows a hierarchical, recursive divide-and-conquer strategy that works as follows: 

\begin{enumerate}
    \item Taking a single cluster with every element, the discretisation technique measures its quality.
    \item If the quality of the cluster does not meet a certain acceptance criterion, the cluster is split into two new smaller clusters. 
    \item Then, the process of measuring and splitting of the clusters is applied to the new clusters. 
\end{enumerate}

This process is repeated recursively until there's no more clusters to split. If, during the process a cluster with a single element is found, the process marks it as outlier, and thus, the algorithm discards it. Cluster splitting is realised using a classical clustering algorithm. This algorithm will try to separate the elements of the cluster in two different groups. This task is realised in PlanMiner-N with the K-means clustering algorithm. 

The K-means algorithm \cite{macqueen1967some} was originally proposed by James MacQueen of the University of California in 1967. From the start, this clustering algorithm stood out in the literature due to its popularity and simplicity. K-means partitions a dataset passed as input into $\mathit{k}$ groups, so that each element is assigned to the group whose mean value is closest.

\begin{algorithm}
\caption{Pseudocode of K-means algorithm}
\label{alg:kmeans}
\textbf{Input} $\mathit{T:}$ Training input data, $\mathit{k:}$ number of desired clusters\\
\textbf{Output} $\mathit{K:}$ Set of clusters
\begin{algorithmic}[1] 
\STATE $\mathit{K} \leftarrow$ Initialize $\mathit{k}$ clusters
\WHILE{$\lnot$ Convergence criterion is met}
\STATE Assign every element of $\mathit{T}$ to the closest cluster in $\mathit{K}$
\STATE Recalculate the mean value of each cluster in $\mathit{K}$
\ENDWHILE
\STATE \textbf{return} $\mathit{K}$ 
\end{algorithmic}
\end{algorithm}

K-means iteratively distributes the different elements $\mathit{T_i}$ of $\mathit{T}$ among the clusters in $\mathit{K}$. In each run of the algorithm (see Algorithm \ref{alg:kmeans}), K-means selects each data point $\mathit{T_i}$ and calculates its distance to the centre of each cluster in $\mathit{K}$, assigning it to the nearest cluster. The centre of each cluster is called the centroid and is calculated as the average value of all the elements assigned to it. In other words, each element $\mathit{T_i}$ is grouped with those elements closest to it. This concept of closeness between two points is computed as the Euclidean distance between them. Once all the data points have been assigned, K-means updates the centroids of each cluster with the information from the new points assigned to them. This process is repeated until the convergence criterion of the algorithm is met, which is that, after one run, all clusters have stabilised. A stable cluster is one where, after one run, no element assigned in the previous run has been added or removed. Finally, it should be noted that the set of clusters $\mathit{K}$ must be initialised at the start before entering the main loop of the algorithm. This can be done randomly (the approach used in this work) or by following a greedy strategy.

The quality of a cluster $\mathit{quality(\textit{cluster})}$ measures how cohesive the values of a cluster are, as well as the size of the clusters. Cluster's quality is defined as the weighted arithmetic mean of both metrics:
$$
quality(C_{i}) = \alpha * silI(C_{i}) + \beta * nSTD(C_{i})
$$
\noindent where $\mathit{silI}$ is the index of the silhouette of the cluster $\mathit{C_{i}}$ \cite{kaufman2009finding}, and $\mathit{nSTD}$ is the normalised standard deviation of $\mathit{C_{i}}$. These metrics, taken from the state-of-the-art of the field in question, and, summarising: 

\begin{itemize}
    \item The \textbf{silhouette index} of a cluster measures how close the elements of a cluster are among themselves and how far they are to the elements of the other clusters. A cluster with a good silhouette index contains well-matched elements which would be bad matches for the elements of the rest of the clusters.
    \item The \textbf{normalised standard deviation} quantifies how ``wide'' a given cluster is. A very wide cluster implies that its elements are dispersed in it; thus, its centre is not representative of them. On the other hand, a very tight cluster would imply that its elements are near the centroid and thus they are well represented by it.
\end{itemize}

On the one hand, the silhouette index of a cluster is computed as the average silhouette index of every data point assigned to it. For a given element $T_i$, its silhouette index is calculated as:
$$s(T_i) =
\begin{cases} 
\frac{b(T_i) - a(T_i)}{max(a(T_i), b(T_i))} \quad & \text{if } |k| > 1\\
0 \quad & \text{if } |k| = 1
\end{cases}
$$
\noindent where $a(T_i)$ is the average distance from the element $T_i$ to the rest of the elements of the cluster:  
$$
a(T_i)={\frac {1}{|k|-1}}\sum _{i,j\in k,i\neq j}distance(T_i,T_j)
$$
\noindent and $b(T_i)$ is the average distance from $\mathit{T_i}$ to the nearest element of the other clusters.
$$
b(T_i)=\min _{k\neq k'}{\frac {1}{|k'|}}\sum _{j\in k'}distance(T_i,T_j)
$$
\noindent with $\mathit{k}$ and $\mathit{k'}$ being elements of the set of clusters $\mathit{K}$.

The $\mathit{distance}$ between two points is calculated by using the Euclidean distance (as in K-Means). The metric $\mathit{s(T_i)}$ ranges from -1 to 1. A value of 1 means that the $\mathit{i}$-th element is perfectly grouped in its assigned cluster; conversely, a value of -1 implies that the element is wrongly classified. The average silhouette score of all points indicates how well grouped the data are in their clusters.

On the other hand, a cluster's normalised standard deviation is calculated as: 
$$nSTD (k) = \frac{\sqrt{\frac{1}{|k|} \sum_{T_{i}\in k}(T_{i} - \mu)^2 }}{\mu}$$

\noindent where $\mathit{k}$ is a cluster of $\mathit{K}$ and $\mu$ its centroid. A cluster's normalised standard deviation quantifies how disperse the elements of the cluster are in relation to the cluster's centroid. The $nSTD (k)$ score has a range of [0,$\infty$). The higher the value of $\mathit{nSTD (k)}$, the wider the cluster, and hence, the more separated the data points are from the centre of the cluster.

These two metrics separately provide useful information about the quality of a cluster, but using them alone would be counterproductive. On the one hand, the silhouette index tends to benefit evenly distributed clusters. For example, when dealing with data that is homogeneously distributed, focusing only on the silhouette index would make our process to promote fit a set of wide clusters covering the entire range of data, but, that would represent poorly the data assigned to them. On the other hand, the single use of the standard deviation would lead to the generation of a large set of very tight clusters, which, taken to the extreme, could generate a cluster for each different value passed as input. These clusters would represent perfectly the datapoint assigned to them but would make the whole discretisation process useless too. Finally, $\alpha$ and $\beta$ are weights for the measures. These weights allow the fine-tuning of the quality function, giving more asymmetric importance to the measures when calculating the quality of a given cluster.

\begin{figure}
    \resizebox{\linewidth}{!}{
    \begin{tikzpicture}
        \begin{axis}[
        scale only axis,
        width=5in,
        height=2in,
        xmin=-4, xmax=6,
        ymin=0, ymax=.7,
        axis on top]
        \addplot[
          ybar,
          bar width=0.035in, 
          bar shift=0in,
          fill=red,
          draw=black] 
          plot coordinates{ 
            (3.25, 0.055) (3, 0.555) (2.97, 0.055) (5, 0.222) (5.05, 0.055) (-3, 0.055)
          };
          
        \addplot[
        ybar,
        bar width=0.23in, 
        bar shift=0.055in,
        fill=blue,
        opacity=0.25,
        draw=black] 
        plot coordinates{ 
         (3, 0.555)
        };
        \node [above] at (axis cs:  3.15,  0.57) {$\mathit{C_1}$};
        
        \addplot[
          ybar,
          bar width=0.1in, 
          bar shift=0.01in,
          fill=green,
          opacity=0.25,
          draw=black] 
          plot coordinates{ 
             (5, 0.222)
          };
          \node [above] at (axis cs:  5, 0.24) {$\mathit{C_2}$};
        
        \end{axis}
        \end{tikzpicture}
    }
    \caption{Example of discretisation. The x-axis shows the different values taken by $\mathit{(bat\_usage\ ?arg1)}$ in the dataset presented in Table \ref{tab:datasetN}, while the y-axis shows the frequency of occurrence of these values. }
    \label{fig:clusters}
\end{figure}
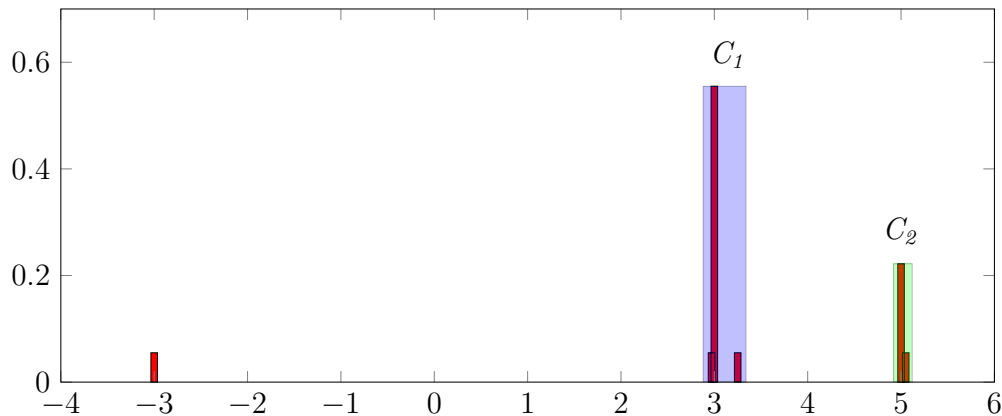

The combination of both metrics allows PlanMiner-N to obtain the best possible clusters, avoiding the problems described above. The main impediment to this combination of metrics is that there are differences between the behaviour of the metrics and their values ranges. Cluster standard deviation is a metric whose scores range in the interval $(0, \infty]$, while the silhouette index is bounded in the interval $[-1, 1]$. In addition, the best values of Cluster standard deviation are the values close to zero, this is opposite to silhouette index where the highest values are the bests scores. Given the behaviour of the metrics, a slightly worse score in the cluster's standard deviation may override a much better silhouette index. In order to lessen this issue and combine both metrics, PlanMiner-N adjusts the silhouette index by taking the opposite of the score obtained, and adding 1 to it. This transformation changes silhouette index results to the interval $[0, 2]$ where zero is the best possible value. This makes both indexes combinable in a single metric whose range is $[0, \infty]$, where Zero is the best result. Figure \ref{fig:clusters} shows the data points and clusters of the $\mathit{(bat\_usage\ ?arg1)}$ fluent. This example presents graphically how the values closer to ``3'' are grouped in the same cluster, while those closer to ``5'' are grouped in another different cluster. The value near ``-3'' is marked as an outlier and therefore is ignored in the next stages of PlanMiner-N's execution.

\subsection{Meta-states refinement} \label{sec:ruleComb}
As we have stated before, when dealing with noisy data, we cannot be sure that after extracting a classification model for a given dataset we will get a unique characterisation for the pre-states and another for the post-states (i.e. a rule set with two rules, one for each meta-state). Given the characteristics of the AML process presented in this paper, it is necessary that this condition is always fulfilled. This is because a deterministic action model has only one rule for pre-state and one for post-state. In order to fulfil this constraint, PlanMiner-N implements this classification model refinement process, which merges a ruleset obtained from a classification algorithm, that characterises a single action, into a a pair of rules (one for the pre-state and other for the post-state). This process is performed before attempting to extract the preconditions and effects of the actions of the domain being learned. In addition, it presents an extra difficulty, as the multiple rules that can be learned by classification algorithms for each action may present inconsistencies between them. This is because the classification algorithms may have adjusted elements of some of the rules to noisy information, detecting this is paramount for the correct learning of the planning domains.

The process implemented in PlanMiner-N to deal with this issue aims to (i) obtain the number of meta-states needed by PlanMiner to function correctly and (ii) detect inconsistencies in them. Prior to the application of said process, PlanMiner-N divides the ruleset in two sets, separating the rules that represent the meta-states of the pre-state from those that model the post-state. It then applies the refinement process to each collection of rules separately, obtaining a single rule for each.

This procedure decomposes the model (the ruleset obtained by the classification algorithm) to evaluate its elements (which are rules) using a statistical method. Once decomposed the rulesets, PlanMiner-N creates a repository with elements obtained, and evaluates them, allowing the detection of strange elements that can be considered noise and detecting inconsistencies between these elements. The Meta-state refinement method implements a set of procedures to study and resolve these issues automatically.

The evaluation method is implemented a as a two-step strategy that:
\begin{enumerate}
    \item An anomaly detection strategy is applied to the elements of the rules' antecedents of a given ruleset, filtering those atypical elements found. 
    \item The remaining elements are studied in order to detect conflict between them, combining them in a single rule in the process.
\end{enumerate}

Finally, to clarify that, in the end, the meta-state refinement method generates only the rules necessary for the learning process to proceed (i.e. two rules). 

\subsubsection{Filtering of irrelevant features}
The main objective of the filtering procedure (Algorithm \ref{alg:FIF}) is to detect those elements of the rule antecedents that have an anomalously low frequency of occurrence. Marking them as noisy and discarding them for the rest of the procedure. Briefly recalling, each rule in the ruleset is defined as a conjunction of features $<attr, val>$ that has a weight associated with it. This weight indicates the percentage of examples of a given class covered by the rule in question.

\begin{algorithm}[tb]
\caption{Irrelevant features filtering Algorithm}
\label{alg:FIF}
\textbf{Input}: rules: Ruleset \\
\textbf{Output}: rules\_feat: Ordered set of features

\begin{algorithmic}[1] 
\STATE \textit{rules\_feat} $\leftarrow \{\}$
\FORALL {Rule \textit{rule} in \textit{rules}}
\FORALL{Features \textit{elem} in the antecedent of \textit{rule}}
\STATE \textit{rules\_feat} $\leftarrow$ \textit{rules\_feat} $\cup$ \textit{elem}
\ENDFOR
\ENDFOR
\STATE Filter(\textit{rules\_feat})
\STATE Short(\textit{rules\_feat})
\STATE \textbf{return} \textit{rules\_feat}
\end{algorithmic}
\end{algorithm}

That said, the implemented in PlanMiner-N to filter irrelevant features of the rules works as follows: Starting from a ruleset that contains several rules, it decompose the antecedents of each rule by separating the features that define them and computing its support value. The support of a feature is calculated by adding the weight of every rule in which it appears. 

Once each feature has been extracted, all those that do not exceed a certain threshold are filtered out. Filter(\textit{rules\_feat}) marks and discards those features that are considered \textit{irrelevant}. The criterion for marking a feature as irrelevant is given by the feature whose support value is higher, which is taken as the reference value. Finally, the filtering process sorts the set of features from highest to lowest according to their support value.

\subsubsection{Conflict resolution}
Once the irrelevant attributes have been filtered and ordered out, the rest of them are combined in a single rule (Algorithm \ref{alg:RL}). This is done by taking the features in descending order and including them in an empty rule incrementally. During this process, PlanMiner-N may find a situation where a feature being included shares the attribute $\mathit{attr}$ with another feature already included in the rule, but with a different value $\mathit{val}$. 

\begin{algorithm}[tb]
\caption{Meta-state Refinement Algorithm}
\label{alg:RL}
\textbf{Input}: rules\_feat: Ordered set of features \\
\textbf{Output}: rule: Rule
\begin{algorithmic}[1] 

\STATE \textit{rule} $\leftarrow \{\}$
\WHILE {\textit{rules\_feat} is not empty}
\STATE \textit{elem} $\leftarrow$ Top feature of  \textit{rules\_feat}
\STATE Delete \textit{elem} from \textit{rules\_feat}
\IF {\textit{stat} does not conflicts with some element of \textit{rule}}
\STATE Add \textit{elem} to \textit{rule} antecedent
\ELSE
\STATE SolveConflict(\textit{rule},\textit{elem})
\ENDIF
\ENDWHILE
\STATE \textbf{return} \textit{rule}
\end{algorithmic}
\end{algorithm}

This is called a \textit{feature conflict} and occurs when the classification algorithms fits a rule to a specific set of data that is inconsistent with other rule. When a conflict is found, PlanMiner-N may face two course of actions: delete the feature with the lowest support (by considering that feature noisy) or delete both features (as considering that the whole feature is irrelevant to model the meta-state). 

SolveConflict(\textit{rule},\textit{elem}) determines said course of action by calculating a confidence interval and checking how the difference between the features' support value \textendash namely $\Delta(\mathit{s_1}, \mathit{s_2})$ \textendash interact with it. The confidence interval is calculated as:

$$[-0.1 *  \bar{S}, 0.1 *  \bar {S}]$$

\noindent where $\mathit{\bar{S}}$ as the mean support value between $\mathit{s_1}$ and $\mathit{s_2}$. 

If $\Delta(\mathit{s_1}, \mathit{s_2})$ is within that interval the difference is considered significant, and then, PlanMiner-N can conclude that the feature with the less support can be discarded. Otherwise, both characteristics are discarded and are not included in the rule that is being built. If no conflict of features arises, the feature is added to the antecedent of the rule being constructed.

\begin{figure}
    \resizebox{\linewidth}{!}{
    \begin{tikzpicture}
    \begin{axis} [xbar,
        bar width = 12pt,
        y = 0.5cm,
        ytick = data,
        tickwidth=0pt,
        enlarge y limits=0.2,
        enlarge x limits=0.1,
        xlabel=Support value,
        ylabel=Predicates,
        symbolic y coords={
        ${(at\ arg1\ arg3) = True}$, ${(at\ arg1\ arg2) = False}$, ${(scanned\ arg3) = False}$, ${(scanned\ arg3) = True}$, ${(at\ arg1\ arg3) = False}$, ${(at\ arg1\ arg2) = True}$, ${(energy\ arg1) > \Delta(energy\ arg1)}$
        },
        nodes near coords,
        nodes near coords style={
            /pgf/number format/precision=4,
        },
        ytick distance=0.01,
        extra x ticks = 1.1,
        extra x tick labels={},
        extra x tick style={grid=major,major grid style={thick,draw=red}}
    ]
    \addplot[
        fill=blue!50,
        opacity=0.75,
        draw=blue] 
        plot coordinates{
            (1,${(at\ arg1\ arg3) = True}$)
            (1,${(at\ arg1\ arg2) = False}$)
            (3,${(scanned\ arg3) = False}$)
            (3,${(scanned\ arg3) = True}$)
            (8,${(at\ arg1\ arg3) = False}$)
            (8,${(at\ arg1\ arg2) = True}$)
            (9,${(energy\ arg1) > \Delta(energy\ arg1)}$)};
    \end{axis}
    \end{tikzpicture}
    }
    \caption{Support values of goto action's classification model.}
    \label{fig:rls}
\end{figure}
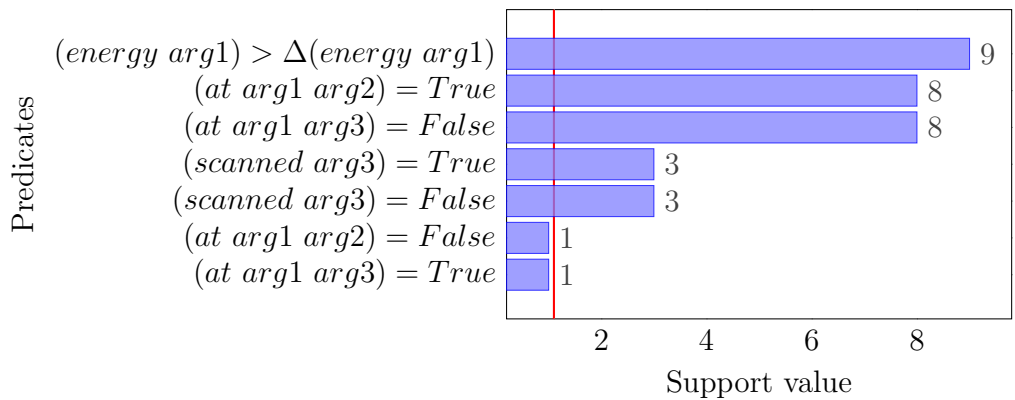

In the example presented in Figure \ref{fig:rls} we can see the support values of the classification rules that describe the pre-states of the $\mathit{(goto\ ?arg1\ ?arg2\ ?arg3)}$ action. Those elements which support value is lesser than the threshold (defined by the red line) are erased. In this example we find a \textit{feature conflict} with the elements $\mathit{(scanned\ arg3) = False}$ and $\mathit{(scanned\ arg3) = True}$, in which both would be erased.

\section{Experimental Setup}
The new methods proposed in this paper were evaluated using a series of planning domains from the International Planning Competition (IPC) \cite{long20033rd}. The evaluation defined in this section tests the capabilities of PlanMiner-N to learn STRIPS and numerical planning domains comparing them with the learning capabilities of the PlanMiner algorithm using noisy data as input. For this purpose, we create a collection of plan traces to be used as input by defining randomly a set of planning problems and solving them. For a single domain, we defined 50 problems and solved them using the FF Planner \cite{hoffmann2003metric}. From these solved problems, a collection of plan traces were extracted to use them as input for the methodologies during the experimentation.

The experimentation is split into two blocks. The first block measures the learning capabilities of the algorithms learning STRIPS planning domains, where the second block aims to test the methodologies by learning planning domains with numerical information. For each block, we use a different set of planning domains. Table \ref{tab:dominios} lists these domains and shows their characteristics.

\begin{table}
    \setlength{\aboverulesep}{-1.0pt}
    \setlength{\belowrulesep}{0.0pt}
    \begin{subtable}[t]{\textwidth}
        \centering
       \begin{tabular}{c|ccccc}
            \toprule
            Domain & $|$Actions$|$ & max action & $|$logical$|$ & max predicate\\
            & & arity & $|$predicates$|$ & arity\\
            \midrule
            BlocksWorld & 4 & 2 & 5 & 2 \\
            Depots & 5 & 4 & 6 & 2 \\
            DriverLog & 6 & 4 & 6 & 2 \\
            ZenoTravel & 5 & 3 & 4 & 2 \\
            \bottomrule
        \end{tabular}
        \caption{STRIPS domains.}
        \label{tab:tablaDoms_a}
    \end{subtable}
    \setlength{\aboverulesep}{-1.0pt}
    \setlength{\belowrulesep}{0.0pt}
    \begin{subtable}[t]{\textwidth}
        \centering
        \begin{tabular}{c|ccccc}
            \toprule
            Domain & $|$Actions$|$  & max action & $|$logical $|$ & $|$numerical$|$ & max predicate\\
            & & arity & $|$predicates$|$ & $|$ predicates$|$ & arity\\
            \midrule
            Depots & 5 & 4 & 6 & 4 & 2 \\
            DriverLog & 6 & 4 & 6 & 4 & 2 \\
            Rovers & 10 & 4 & 26  & 2 & 3 \\
            Satellite & 5 & 4 & 8 & 6 & 2 \\
            ZenoTravel & 5 & 3 & 2 & 8 & 2 \\
            \bottomrule
        \end{tabular}
        \caption{Numerical domains.}
        \label{tab:tablaDoms_b}
    \end{subtable}
    \caption{Input domains characteristics. This Table shows the domains used in the second part of the experimentation. Domains characteristics (from left to right) are the number of actions of the domains, the number of parameters of the actions, the number of logic predicates, the number of numerical predicates (if any), and the maximum number of parameters of the predicates.}
    \label{tab:dominios}
\end{table}

As the plan traces obtained from a set of solved planning problems are noise-free, in order to experiment with the capabilities of the approaches learning from noisy input information, artificial noise is included in them. The noise was included by selecting randomly a certain percentage of the predicates of the plan traces' states and changing their values. Noise inclusion follows two different strategies: a) outliers inclusion ie. change the values of the logical predicates and the numerical predicates by arbitrary numbers, and b) modifying the values of the numerical predicates with random noise ie. making them non-CRISP. In each experimentation block, we have defined several different experiments by applying different types of noise to the plan traces. When learning STRIPS-only planning domains, we only include outliers in the plan traces. On the other hand, when learning numerical domains, we define three different experiments: i) including outliers, ii) including CRISP values, and iii) including both outliers and CRISP values. Finally, tolerance to noise is tested in the experimentation by defining a set of the experiments with different percentages of noise in the input plan traces and obtaining a planning domain from them.

From these learned domains, we measure certain aspects of them in order to test the learning capabilities of the algorithms. The measures evaluate the quality of the learned domains and, therefore, the quality of the methodologies used to obtain them. The quality measures used to evaluate the results of the new methods are precision, recall, F-Score \cite{AINETO2019104} and validity \cite{valsprogress}. Precision and recall measure the number of missing and surplus elements in the preconditions and effects of the measured domain's actions. For a given action in a domain, its precision is measured as: $$Precision = \frac{tp}{tp + fp}$$ and its recall as: $$Recall = \frac{tp}{tp + fn}$$ where \textit{tp} is is the number of true positives (correct elements in its preconditions and effects), \textit{fn} is the number of false negatives (missing elements in its preconditions and effects), and \textit{fp} is the number of false positives (extra elements in its preconditions and effects). F-Score is a measure of the action's overall quality, and is defined as the harmonic mean between precision and recall. F-Score is calculated as: $$F\mbox{-}Score = 2 * \frac{Precision*Recall}{Precision+Recall}$$ The quality of a planning domain is calculated as the average quality measured of every action contained in it.

The validation of a planning domain measures the ability of the domain to solve planning problems. This is calculated by trying to solve planning problems with the learned domain, and checking if the obtained plans are correct. To check if a plan obtained is correct, we start from the initial state of the problem and apply the actions of the plan until we obtain the final state. Then, if the final state is equal to the objective state, then the plan is correct. A domain is valid if it is able to obtain valid plans for all the problems in a set of validation problems. The task of validating a plan is performed with the VAL tool used in the IPC.

Finally, in order to present reliable results and avoid the influence of the randomness of the inclusion of noise, each different experiment is carried on using a 5 fold cross-validation technique. Using this technique, we split the input plan traces into five subsets (or folds), selecting four of them as training data and the remaining set as test data. The plan traces in the training data are used to learn planning domains, where the plan traces in the test data are used to validate them. Each experiment is executed five times, selecting a different fold as test data in each run. The results of the experiment are the average result of every run.

\subsection{Algorithms used}
During the experimental process, different versions of PlanMiner and PlanMiner-N have been evaluated, where each one of the versions uses a different algorithm for generating the classification models. The aim of this is to test the feasibility and robustness of the learning pipeline regardless of the learning engine used in it. The classification algorithms used in this experimentation are ID3 \cite{quinlan1986induction}, C4.5 \cite{quinlan2014c4}, RIPPER \cite{cohen1995fast} and NSLV \cite{10.1016/j.ijar.2015.09.001}. A full description of these algorithms can be found in the referenced papers.

In addition to classification algorithms, a set of state-of-the-art AML algorithms was selected as reference algorithms for the experimental process. These algorithms are ARMS \cite{Yang07}, AMAN \cite{zhuo2013refining}, OpMaker2 \cite{mccluskey2009automated} and FAMA \cite{AINETO2019104}. Due to the limitations of these algorithms for learning planning domains with numerical information, they are only used during experimentation with STRIPS domains.

In these experiments, the parameters of PlanMiner, PlanMiner-N, FF-Metric, VAL and the reference algorithms are set as default as noted by its authors in its reference works. Table \ref{tab:Exppar} displays these parameters and their impact on a given algorithm’s performance. If an algorithm does not appear in the mentioned table it is because it has not any parameter to set before its execution. With regard to the different parameter settings, we include a brief description of them and their impact on the performance of the given algorithms (except for PlanMiner-N, as this information has already been given in the relevant sections above). PlanMiner-N uses the same parameter settings as PlanMiner in addition to its particular parameters setting. That said, for each algorithm the parameter's settings are:

\begin{itemize}
    \item \textbf{Metric-FF parameters setting.} MetricFF's parameters weight the components of the heuristic that governs the internal search process of the planner. This weights change the importance that MetricFF gives to the estimated cost to the goal (\textit{h}) and the cost of the current path explored (\textit{g}) when guiding the search process. These parameters have been set as 1 to avoid interfere with the search process, giving the same weight to both elements of the heuristic.
    
    \item \textbf{ARMS parameters setting.} The \textit{probability threshold} of ARMS is used to filter which information contained in the input plan traces is considered as a constraint when building the logic formulas used to learn action models. The lower the threshold, more information is considered. This increases the computation time of ARMS, but it may consider in the learning process information with a low appearance rate in the input data, but that can be useful. The probability threshold value (0.7) is recommended by the authors of ARMS as is the best value that balances the amount of information processed and the results of the algorithm.
    
    \item \textbf{AMAN parameters setting.} AMAN builds iteratively a set of action models from a partial set data extracted from the input plan traces. Increasing the number of \textit{iterations} leads to an increase in the number of the set of action models generated. This increases the probability of finding the best set of action models possible at the cost of a bigger computation time. The authors recommend the given number of iterations (1500 iterations) as they are the most efficient way to obtain good results.
\end{itemize}

\begin{table}
    \setlength{\aboverulesep}{-0.5pt}
    \setlength{\belowrulesep}{0.0pt}
    \centering
    \begin{tabular}{c|c}
    \toprule
    \textbf{PlanMiner parameters} & Value\\ 
    Symbolic Regression acceptance threshold & 0.02\\
    Symbolic Regression timeout & 300\\
    \midrule
    \textbf{PlanMiner-N parameters} & Value\\ 
    Statistical noise filtering threshold & 5\%\\
    Cluster's quality \textit{alpha} & 0.6\\
    Cluster's quality \textit{beta} & 0.4\\
    Cluster's acceptance criterion & 0.05\\
    Irrelevant features detection threshold & 0.05\\
    \midrule
    \textbf{MetricFF parameters} & \\ 
    H weight &1\\
    G weight &1\\
    \midrule
    \textbf{ARMS parameters} & \\ 
    Probability threshold & 0.7\\
    \midrule
    \textbf{AMAN parameters} & \\ 
    Number of iterations & 1500\\
    \bottomrule
    \end{tabular}
    \caption{Settings of the different algorithms during the experimentation process.}
    \label{tab:Exppar}
\end{table}

\subsection{Learning STRIPS domains}
\subsubsection{Classification algorithms comparison.} 
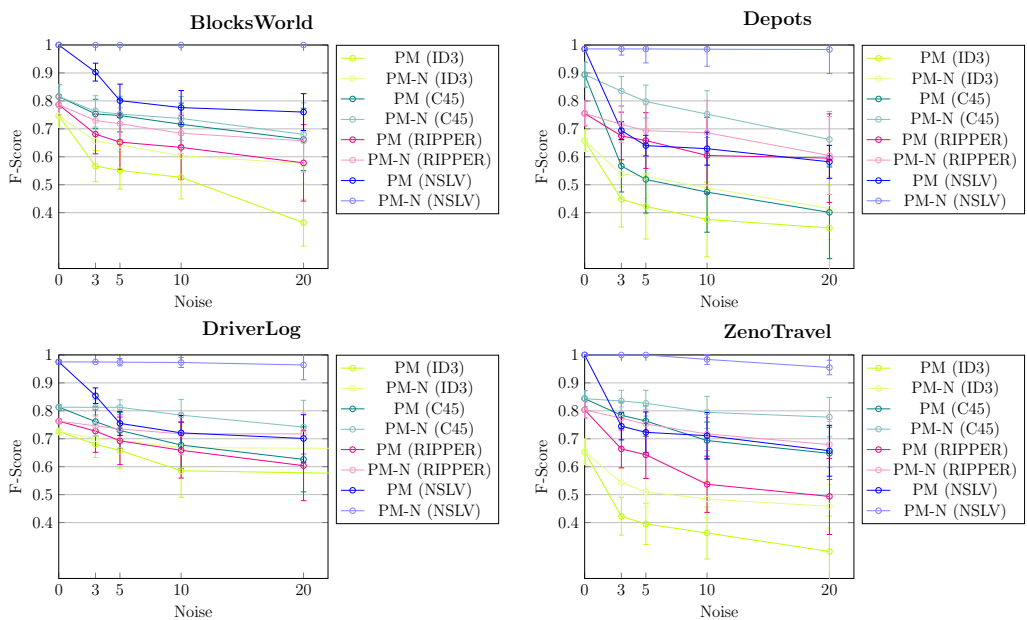
\begin{figure}
    \centering
    \resizebox{0.49\linewidth}{!}{
            \begin{tikzpicture}
                \begin{axis}[
                    ylabel={F-Score},
                    xlabel={Noise},
                    ymin=0.2,
                    ymax=1.0,
                    xmin=0,
                    xmax=22,
                    ytick={0.4, 0.5, 0.6, 0.7, 0.8, 0.9, 1.0},
                    xtick={0, 3, 5, 10, 20},
                    legend pos=outer north east,
                    ymajorgrids=true,
                ]
               \addplot[
                    color=lime,
                    mark=o,
                    error bars/.cd,
                    y dir=both,
                    y explicit
                    ]
                    coordinates {
                        (0, 0.7463)+=(0, 0.0414)-=(0, 0.0414)
                        (3, 0.5667)+=(3, 0.0560)-=(3, 0.0560)
                        (5, 0.5510)+=(5, 0.0666)-=(5, 0.0666)
                        (10, 0.5263)+=(10, 0.0772)-=(10, 0.0772)
                        (20, 0.3642)+=(20, 0.0837)-=(20, 0.0837)
                    };
                \addlegendentry{PM (ID3)};
                
               \addplot[
                    color=lime!50,
                    mark=o,
                    error bars/.cd,
                    y dir=both,
                    y explicit
                    ]
                    coordinates {
                        (0, 0.7463)+=(0, 0.0414)-=(0, 0.0414)
                        (3, 0.6562)+=(3, 0.0432)-=(3, 0.0432)
                        (5, 0.6425)+=(5, 0.0437)-=(5, 0.0437)
                        (10, 0.6039)+=(10, 0.0721)-=(10, 0.0721)
                        (20, 0.5758)+=(20, 0.0728)-=(20, 0.0728)
                    };
                \addlegendentry{PM-N (ID3)};

                \addplot[
                    color=teal,
                    mark=o,
                    error bars/.cd,
                    y dir=both,
                    y explicit
                    ]
                    coordinates {
                        (0, 0.8157)+=(0, 0.0425)-=(0, 0.0425)
                        (3, 0.7532)+=(3, 0.0528)-=(3, 0.0528)
                        (5, 0.7476)+=(5, 0.0582)-=(5, 0.0582)
                        (10, 0.7163)+=(10, 0.0757)-=(10, 0.0757)
                        (20, 0.6620)+=(20, 0.1114)-=(20, 0.1114)
                    };
                \addlegendentry{PM (C45)};
                \addplot[
                    color=teal!50,
                    mark=o,
                    error bars/.cd,
                    y dir=both,
                    y explicit
                    ]
                    coordinates {
                        (0, 0.8157)+=(0, 0.0425)-=(0, 0.0425)
                        (3, 0.7624)+=(3, 0.0561)-=(3, 0.0561)
                        (5, 0.7526)+=(5, 0.0638)-=(5, 0.0638)
                        (10, 0.7372)+=(10, 0.0782)-=(10, 0.0782)
                        (20, 0.6803)+=(20, 0.1127)-=(20, 0.1127)
                    };
                \addlegendentry{PM-N (C45)};

                \addplot[
                    color=magenta,
                    mark=o,
                    error bars/.cd,
                    y dir=both,
                    y explicit
                    ]
                    coordinates {
                        (0, 0.7863)+=(0, 0.0361)-=(0, 0.0361)
                        (3, 0.6811)+=(3, 0.0706)-=(3, 0.0706)
                        (5, 0.6524)+=(5, 0.0972)-=(5, 0.0972)
                        (10, 0.6336)+=(10, 0.1152)-=(10, 0.1152)
                        (20, 0.5782)+=(20, 0.1367)-=(20, 0.1367)
                    };
                \addlegendentry{PM (RIPPER)};
                \addplot[
                    color=magenta!50,
                    mark=o,
                    error bars/.cd,
                    y dir=both,
                    y explicit
                    ]
                    coordinates {
                        (0, 0.7863)+=(0, 0.0361)-=(0, 0.0361)
                        (3, 0.7300)+=(3, 0.0441)-=(3, 0.0441)
                        (5, 0.7182)+=(5, 0.0571)-=(5, 0.0571)
                        (10, 0.6843)+=(10, 0.0725)-=(10, 0.0725)
                        (20, 0.6571)+=(20, 0.1117)-=(20, 0.1117)
                    };
                \addlegendentry{PM-N (RIPPER)};

                \addplot[
                    color=blue,
                    mark=o,
                    error bars/.cd,
                    y dir=both,
                    y explicit
                    ]
                    coordinates {
                        (0, 1.000)+=(0, 0)-=(0, 0)
                        (3, 0.903)+=(3, 0.032)-=(3, 0.032)
                        (5, 0.801)+=(5, 0.059)-=(5, 0.059)
                        (10, 0.776)+=(10, 0.061)-=(10, 0.061)
                        (20, 0.760)+=(20, 0.066)-=(20, 0.066)
                    };
                \addlegendentry{PM (NSLV)};
                
                \addplot[
                    color=blue!50,
                    mark=o,
                    error bars/.cd,
                    y dir=both,
                    y explicit
                    ]
                    coordinates {
                        (0, 1)+=(0, 0)-=(0, 0)
                        (3, 1)+=(3, 0)-=(3, 0)
                        (5, 1)+=(5, 0)-=(5, 0)
                        (10, 1)+=(10, 0)-=(10, 0)
                        (20, 1)+=(20, 0)-=(20, 0)
                    };
                \addlegendentry{PM-N (NSLV)};
                
                \end{axis},
            \node[above,font=\large\bfseries] at (current bounding box.north) {BlocksWorld};
            \end{tikzpicture}
    }    
    \hspace{0mm}
    \resizebox{0.49\linewidth}{!}{
            \begin{tikzpicture}
                \begin{axis}[
                    ylabel={F-Score},
                    xlabel={Noise},
                    ymin=0.2,
                    ymax=1.0,
                    xmin=0,
                    xmax=22,
                    ytick={0.4, 0.5, 0.6, 0.7, 0.8, 0.9, 1.0},
                    xtick={0, 3, 5, 10, 20},
                    legend pos=outer north east,
                    ymajorgrids=true,
                ]
                
                \addplot[
                    color=lime,
                    mark=o,
                    error bars/.cd,
                    y dir=both,
                    y explicit
                    ]
                    coordinates {
                        
                        (0, 0.6574)+=(0,0.0382)-=(0, 0.0382)
                        (3, 0.4477)+=(3, 0.0992)-=(3, 0.0992)
                        (5, 0.4223)+=(5, 0.1166)-=(5, 0.1166)
                        (10, 0.3757)+=(10, 0.1336)-=(10, 0.1336)
                        (20, 0.3452)+=(20, 0.1535)-=(20, 0.1535)
                    };
                \addlegendentry{PM (ID3)};
                \addplot[
                    color=lime!50,
                    mark=o,
                    error bars/.cd,
                    y dir=both,
                    y explicit
                    ]
                    coordinates {
                        (0, 0.6574)+=(0,0.0382)-=(0, 0.0382)
                        (3, 0.5383)+=(3, 0.0472)-=(3, 0.0472)
                        (5, 0.5293)+=(5, 0.0783)-=(5, 0.0783)
                        (10, 0.4887)+=(10, 0.0892)-=(10, 0.0892)
                        (20, 0.4154)+=(20, 0.1119)-=(20, 0.1119)
                    };
                \addlegendentry{PM-N (ID3)};

                \addplot[
                    color=teal,
                    mark=o,
                    error bars/.cd,
                    y dir=both,
                    y explicit
                    ]
                    coordinates {
                        (0, 0.8939)+=(0,0.0445)-=(0, 0.0445)
                        (3, 0.5672)+=(3, 0.0932)-=(3, 0.0932)
                        (5, 0.5189)+=(5, 0.1202)-=(5, 0.1202)
                        (10, 0.4736)+=(10, 0.1435)-=(10, 0.1435)
                        (20, 0.4010)+=(20, 0.1656)-=(20, 0.1656)
                    };
                \addlegendentry{PM (C45)};
                \addplot[
                    color=teal!50,
                    mark=o,
                    error bars/.cd,
                    y dir=both,
                    y explicit
                    ]
                    coordinates {
                        (0, 0.8939)+=(0,0.0445)-=(0, 0.0445)
                        (3, 0.8352)+=(3, 0.0523)-=(3, 0.0523)
                        (5, 0.7978)+=(5, 0.0589)-=(5, 0.0589)
                        (10, 0.7527)+=(10, 0.0836)-=(10, 0.0836)
                        (20, 0.6620)+=(20, 0.1003)-=(20, 0.1003)
                    };
                \addlegendentry{PM-N (C45)};

                \addplot[
                    color=magenta,
                    mark=o,
                    error bars/.cd,
                    y dir=both,
                    y explicit
                    ]
                    coordinates {
                        (0, 0.7553)+=(0,0.0445)-=(0, 0.0445)
                        (3, 0.6747)+=(3, 0.0852)-=(3, 0.0852)
                        (5, 0.6582)+=(5, 0.0997)-=(5, 0.0997)
                        (10, 0.6047)+=(10, 0.1367)-=(10, 0.1367)
                        (20, 0.5952)+=(90, 0.1584)-=(90, 0.1584)
                    };
                \addlegendentry{PM (RIPPER)};
                \addplot[
                    color=magenta!50,
                    mark=o,
                    error bars/.cd,
                    y dir=both,
                    y explicit
                    ]
                    coordinates {
                        (0, 0.7553)+=(0,0.0445)-=(0, 0.0445)
                        (3, 0.7157)+=(3, 0.0639)-=(3, 0.0639)
                        (5, 0.6935)+=(5, 0.0925)-=(5, 0.0925)
                        (10, 0.6861)+=(10, 0.1159)-=(10, 0.1159)
                        (20, 0.6046)+=(90, 0.1396)-=(90, 0.1396)
                    };
                \addlegendentry{PM-N (RIPPER)};

                \addplot[
                    color=blue,
                    mark=o,
                    error bars/.cd,
                    y dir=both,
                    y explicit
                    ]
                    coordinates {
                        (0, 0.986)+=(0, 0)-=(0, 0)
                        (3, 0.694)+=(3, 0.032)-=(3, 0.032)
                        (5, 0.640)+=(5, 0.037)-=(5, 0.037)
                        (10, 0.629)+=(10, 0.059)-=(10, 0.059)
                        (20, 0.582)+=(20, 0.059)-=(20, 0.059)
                    };

                \addlegendentry{PM (NSLV)};
                \addplot[
                    color=blue!50,
                    mark=o,
                    error bars/.cd,
                    y dir=both,
                    y explicit
                    ]
                    coordinates {
                        (0, 0.986)+=(0, 0)-=(0, 0)
                        (3, 0.986)+=(3, 0.022)-=(3, 0.022)
                        (5, 0.986)+=(5, 0.050)-=(5, 0.050)
                        (10, 0.985)+=(10, 0.061)-=(10, 0.061)
                        (20, 0.984)+=(20, 0.086)-=(20, 0.086)
                    };
                    
                \addlegendentry{PM-N (NSLV)};
                \end{axis},
            \node[above,font=\large\bfseries] at (current bounding box.north) {Depots};
            \end{tikzpicture}
    }   
    
    \resizebox{0.49\linewidth}{!}{
            \begin{tikzpicture}
                \begin{axis}[
                    ylabel={F-Score},
                    xlabel={Noise},
                    ymin=0.2,
                    ymax=1.0,
                    xmin=0,
                    xmax=22,
                    ytick={0.4, 0.5, 0.6, 0.7, 0.8, 0.9, 1.0},
                    xtick={0, 3, 5, 10, 20},
                    legend pos=outer north east,
                    ymajorgrids=true,
                ]
    
                \addplot[
                    color=lime,
                    mark=o,
                    error bars/.cd,
                    y dir=both,
                    y explicit
                    ]
                    coordinates {
                        (0, 0.7259)+=(0,0.0115)-=(0, 0.0115)
                        (3, 0.6802)+=(3, 0.0472)-=(3, 0.0472)
                        (5, 0.6582)+=(5, 0.0628)-=(5, 0.0628)
                        (10, 0.5853)+=(10, 0.0945)-=(10, 0.0945)
                        (90, 0.5287)+=(20, 0.1125)-=(20, 0.1125)
                    };
                \addlegendentry{PM (ID3)};
                \addplot[
                    color=lime!50,
                    mark=o,
                    error bars/.cd,
                    y dir=both,
                    y explicit
                    ]
                    coordinates {
                        (0, 0.7259)+=(0,0.0115)-=(0, 0.0115)
                        (3, 0.7003)+=(3, 0.0297)-=(3, 0.0297)
                        (5, 0.6945)+=(5, 0.0461)-=(5, 0.0461)
                        (10, 0.6680)+=(10, 0.0737)-=(10, 0.0737)
                        (90, 0.6537)+=(20, 0.0855)-=(20, 0.0855)
                    };
                \addlegendentry{PM-N (ID3)}; 
                \addplot[
                    color=teal,
                    mark=o,
                    error bars/.cd,
                    y dir=both,
                    y explicit
                    ]
                    coordinates {
                        (0, 0.8130)+=(0,0.0089)-=(0, 0.0089)
                        (3, 0.7614)+=(3, 0.0528)-=(3, 0.0528)
                        (5, 0.7298)+=(5, 0.0638)-=(5, 0.0638)
                        (10, 0.6774)+=(10, 0.0836)-=(10, 0.0836)
                        (20, 0.6262)+=(20, 0.1163)-=(20, 0.1163)                        
                    };
                \addlegendentry{PM (C45)};
                \addplot[
                    color=teal!50,
                    mark=o,
                    error bars/.cd,
                    y dir=both,
                    y explicit
                    ]
                    coordinates {
                        (0, 0.8130)+=(0,0.0089)-=(0, 0.0089)
                        (3, 0.8125)+=(3, 0.0270)-=(3, 0.0270)
                        (5, 0.8125)+=(5, 0.0270)-=(5, 0.0270)
                        (10, 0.7845)+=(10, 0.0563)-=(10, 0.0563)
                        (20, 0.7418)+=(20, 0.0957)-=(20, 0.0957)
                    };
                \addlegendentry{PM-N (C45)};
                
                \addplot[
                    color=magenta,
                    mark=o,
                    error bars/.cd,
                    y dir=both,
                    y explicit
                    ]
                    coordinates {
                        (0, 0.7630)+=(0,0.0089)-=(0, 0.0089)
                        (3, 0.7273)+=(3, 0.0759)-=(3, 0.0759)
                        (5, 0.6926)+=(5, 0.0852)-=(5, 0.0852)
                        (10, 0.6587)+=(10, 0.0994)-=(10, 0.0994)
                        (20, 0.6036)+=(20, 0.1253)-=(20, 0.1253)
                    };
                \addlegendentry{PM (RIPPER)};
                \addplot[
                    color=magenta!50,
                    mark=o,
                    error bars/.cd,
                    y dir=both,
                    y explicit
                    ]
                    coordinates {
                        (0, 0.7630)+=(0,0.0089)-=(0, 0.0089)
                        (3, 0.7462)+=(3, 0.0350)-=(3, 0.0350)
                        (5, 0.7351)+=(5, 0.0512)-=(5, 0.0512)
                        (10, 0.7175)+=(10, 0.0746)-=(10, 0.0746)
                        (20, 0.7175)+=(20, 0.0746)-=(20, 0.0746)
                    };
                \addlegendentry{PM-N (RIPPER)};

                \addplot[
                    color=blue,
                    mark=o,
                    error bars/.cd,
                    y dir=both,
                    y explicit
                    ]
                    coordinates {
                        (0, 0.975)+=(0, 0)-=(0, 0)
                        (3, 0.854)+=(3, 0.028)-=(3, 0.028)
                        (5, 0.755)+=(5, 0.043)-=(5, 0.043)
                        (10, 0.721)+=(10, 0.062)-=(10, 0.062)
                        (20, 0.701)+=(20, 0.085)-=(20, 0.085)
                    };
                \addlegendentry{PM (NSLV)};
                
                \addplot[
                    color=blue!50,
                    mark=o,
                    error bars/.cd,
                    y dir=both,
                    y explicit
                    ]
                    coordinates {
                        (0, 0.975)+=(0, 0)-=(0, 0)
                        (3, 0.975)+=(3, 0.000)-=(3, 0.000)
                        (5, 0.974)+=(5, 0.013)-=(5, 0.013)
                        (10, 0.973)+=(10, 0.018)-=(10, 0.018)
                        (20, 0.964)+=(20, 0.053)-=(20, 0.053)
                    };
                    
                \addlegendentry{PM-N (NSLV)};
                \end{axis},
            \node[above,font=\large\bfseries] at (current bounding box.north) {DriverLog};
            \end{tikzpicture}
    }    
    \hspace{0mm}
    \resizebox{0.49\linewidth}{!}{
            \begin{tikzpicture}
                \begin{axis}[
                    ylabel={F-Score},
                    xlabel={Noise},
                    ymin=0.2,
                    ymax=1.0,
                    xmin=0,
                    xmax=22,
                    ytick={0.4, 0.5, 0.6, 0.7, 0.8, 0.9, 1.0},
                    xtick={0, 3, 5, 10, 20},
                    legend pos=outer north east,
                    ymajorgrids=true,
                ]
                
                \addplot[
                    color=lime,
                    mark=o,
                    error bars/.cd,
                    y dir=both,
                    y explicit
                    ]
                    coordinates {
                        (0, 0.6533)+=(0,0.0456)-=(0, 0.0456)
                        (3, 0.4224)+=(3, 0.0674)-=(3, 0.0674)
                        (5, 0.3952)+=(5, 0.0739)-=(5, 0.0739)
                        (10, 0.3627)+=(10, 0.0932)-=(10, 0.0932)
                        (20, 0.2967)+=(20, 0.1264)-=(20, 0.1264)
                    };
                \addlegendentry{PM (ID3)};  
                \addplot[
                    color=lime!50,
                    mark=o,
                    error bars/.cd,
                    y dir=both,
                    y explicit
                    ]
                    coordinates {
                        (0, 0.6533)+=(0,0.0456)-=(0, 0.0456)
                        (3, 0.5432)+=(3, 0.0509)-=(3, 0.0509)
                        (5, 0.5101)+=(5, 0.0535)-=(5, 0.0535)
                        (10, 0.4842)+=(10, 0.0682)-=(10, 0.0682)
                        (20, 0.4582)+=(20, 0.0783)-=(20, 0.0783)
                    };
                \addlegendentry{PM-N (ID3)};  
                
                \addplot[
                    color=teal,
                    mark=o,
                    error bars/.cd,
                    y dir=both,
                    y explicit
                    ]
                    coordinates {
                        (0, 0.8435)+=(0,0.0287)-=(0, 0.0287)
                        (3, 0.7842)+=(3, 0.0471)-=(3, 0.0471)
                        (5, 0.7634)+=(5, 0.0545)-=(5, 0.0545)
                        (10, 0.6936)+=(10, 0.0663)-=(10, 0.0663)
                        (20, 0.6478)+=(20, 0.0934)-=(20, 0.0934)
                    };
                \addlegendentry{PM (C45)};
                \addplot[
                    color=teal!50,
                    mark=o,
                    error bars/.cd,
                    y dir=both,
                    y explicit
                    ]
                    coordinates {
                        (0, 0.8435)+=(0,0.0287)-=(0, 0.0287)
                        (3, 0.8352)+=(3, 0.0385)-=(3, 0.0385)
                        (5, 0.8272)+=(5, 0.0461)-=(5, 0.0461)
                        (10, 0.7946)+=(10, 0.0572)-=(10, 0.0572)
                        (20, 0.7773)+=(20, 0.0703)-=(20, 0.0703)
                    };
                \addlegendentry{PM-N (C45)};
                
                \addplot[
                    color=magenta,
                    mark=o,
                    error bars/.cd,
                    y dir=both,
                    y explicit
                    ]
                    coordinates {
                        (0, 0.8035)+=(0,0.0287)-=(0, 0.0287)
                        (3, 0.6637)+=(3, 0.0678)-=(3, 0.0678)
                        (5, 0.6426)+=(5, 0.0849)-=(5, 0.0849)
                        (10, 0.5371)+=(10, 0.1006)-=(10, 0.1006)
                        (20, 0.4936)+=(20, 0.1363)-=(20, 0.1363)
                    };
                \addlegendentry{PM (RIPPER)};
                \addplot[
                    color=magenta!50,
                    mark=o,
                    error bars/.cd,
                    y dir=both,
                    y explicit
                    ]
                    coordinates {
                        (0, 0.8035)+=(0,0.0287)-=(0, 0.0287)
                        (3, 0.7736)+=(3, 0.0402)-=(3, 0.0402)
                        (5, 0.7528)+=(5, 0.0526)-=(5, 0.0526)
                        (10, 0.7172)+=(10, 0.0588)-=(10, 0.0588)
                        (20, 0.6792)+=(20, 0.0821)-=(20, 0.0821)
                    };
                \addlegendentry{PM-N (RIPPER)};
                
                \addplot[
                    color=blue,
                    mark=o,
                    error bars/.cd,
                    y dir=both,
                    y explicit
                    ]
                    coordinates {
                        (0, 1.000)+=(0, 0)-=(0, 0)
                        (3, 0.745)+=(3, 0.049)-=(3, 0.049)
                        (5, 0.723)+=(5, 0.073)-=(5, 0.073)
                        (10, 0.711)+=(10, 0.083)-=(10, 0.083)
                        (20, 0.657)+=(20, 0.091)-=(20, 0.091)
                    };
                \addlegendentry{PM (NSLV)};
                \addplot[
                    color=blue!50,
                    mark=o,
                    error bars/.cd,
                    y dir=both,
                    y explicit
                    ]
                    coordinates {
                        (0, 1.000)+=(0, 0)-=(0, 0)
                        (3, 1.000)+=(3, 0.000)-=(3, 0.000)
                        (5, 1.000)+=(5, 0.000)-=(5, 0.000)
                        (10, 0.984)+=(10, 0.018)-=(10, 0.018)
                        (20, 0.955)+=(20, 0.026)-=(20, 0.026)
                    };
                    
                \addlegendentry{PM-N (NSLV)};
                
                \end{axis},
            \node[above,font=\large\bfseries] at (current bounding box.north) {ZenoTravel};
            \end{tikzpicture}
    }   
    \caption{Performance comparison of PlanMiner-N using different classification algorithms on STRIPS domains.}
    \label{fig:compC4STRIPSPM}
\end{figure}

The experimental process begins by studying how the selected classification algorithm affects the performance of PlanMiner. Figure \ref{fig:compC4STRIPSPM} shows these performances in terms of F-Score, while Table \ref{tab:validitySTRIPSC4PM} shows the validity results of the battery of experiments. The Figure \ref{fig:compC4STRIPSPM} fixes in X-axis the incompleteness degree of the plan traces, and Tables are shorted by domains in order to improve their readability. For the sake of readibility, this section presents a summary version of the results, showing only the F-Score and Validity metrics of the domains learned. 

If we look closely at the results the figure \ref{fig:compC4STRIPSPM} we can see the following:
\begin{itemize}
    \item \textbf{BlocksWorld.} PlanMiner-N (NSLV) shows perfect results throughout the whole experimentation and is impervious to the effects of noise. Compared to PlanMiner (NSLV), PlanMiner-N (NSLV) shows almost 25\% higher F-Score in the noisiest experiment. These differences also occur between PlanMiner-N (ID3) and PlanMiner (ID3), PlanMiner-N (C45) and PlanMiner (C45), and between PlanMiner-N (RIPPER) and PlanMiner (RIPPER), with an average variation of 10\% throughout the experimentation. In the most complex experiments, NSLV performs better than ID3, C45 and RIPPER. 
    \item \textbf{Depots.} As with BlocksWorld, PlanMiner-N (NSLV) has a high resistance to noise, which is unchanged throughout the experimentation. PlanMiner (NSLV), on the other hand, suffers a severe drop in performance when some noise is included, which puts its F-Score almost 30 points below PlanMiner-N. The rest of the algorithms have similar behaviour to PlanMiner (NSLV), suffering from a performance drop when noise is included in the plan traces. PlanMiner-N shows values around 15-20\% higher than PlanMiner using the same classification algorithm. 
    \item \textbf{DriverLog.} The difference between PlanMiner-N (NSLV) and PlanMiner (NSLV) when dealing with some noise is 15\% F-Score, a difference that increases to 30\% using data with 20\% noisy elements. PlanMiner-N (ID3) and PlanMiner(ID3) suffer an initial drop in performance (more pronounced in PlanMiner), but then stabilise and remain unchanged even at the highest noise levels. Using the C45 and RIPPER classification algorithms, similar behaviour to NSLV is observed, but not as marked with PlanMiner-N showing unchanged in the initial experiments, but dropping slightly in the final ones. Even so, the drops are much smaller than those seen with PlanMiner.
    \item \textbf{ZenoTravel.} PlanMiner-N (NSLV) obtains perfect results until it encounters 10\% and 20\% noise, where it drops to 98\% and 96\% F-Score respectively. PlanMiner (NSLV) drops from 100\% F-Score to 74\% when encountering some noise, this drop continues (although somewhat more controlled) throughout the experimentation, presenting results below 77\% F-Score in the more complex experimental assumptions. The results of PlanMiner-N with ID3, C45 and RIPPER behave similarly to PlanMiner-N (NSLV), showing some noise resilience compared to their PlanMiner counterparts which lose a lot of performance when encountering some noise.
\end{itemize}

In general, all algorithms exhibit identical behaviour with a large drop in performance when noise is included in the plan traces with the exception of PlanMiner-n (NSLV). Since the PlanMiner algorithm is not designed to work with noisy information, severe performance losses are to be expected for it. On the other hand, PlanMiner-N, the approach tested in this experimentation, presents far better results than PlanMiner as expected too. PlanMiner-N shows some resistance to noise, but, its performance is highly dependent on the classification algorithm used in the learning pipeline. This difference in the performance of PlanMiner-N can be seen in that, while experiments performed with NSLV remain somewhat stable throughout the experiment, those using the ID3 classifier, for example, show a drop of 10-12 points when noise is included in the plan strokes. This is due to the fact that, because of NSLV's ability to obtain descriptive rules from the datasets (i.e. rules that contain all the information necessary to represent a set of examples), the algorithm is able to generate sets of rules that fully explain the input information (including noisy examples). What is a priori a loss of generalisation of the algorithm is a blessing for PlanMiner-N, as it gives it extensive knowledge about the data, helping it to filter out the noise in the data. The other algorithms do not enjoy this advantage, providing less information to the learning algorithm. This means that PlanMiner is not able to correctly refine the models obtained with them. Nevertheless, if we compare PlanMiner and PlanMiner-N using the same classifier, we see the clear superiority of the latter over the former regardless of the classification algorithm used. Using the ID3 classification algorithm, PlanMiner shows F-Score values below 30\% in some experiments, while PlanMiner-N obtains 15 points more F-Score in the same experiments. With C.45 and RIPPER, PlanMiner-N shows an improvement of around 20\% over PlanMiner throughout the experimental process. Although the biggest difference in performance can be observed with NSLV, since while PlanMiner obtains 58\% F-Score results, PlanMiner-N obtains perfect results. This indicates that the changes made in PlanMiner-N are effective in addressing the learning problem using noisy input data. 

\begin{table}
\setlength{\aboverulesep}{-0.5pt}
\setlength{\belowrulesep}{0.0pt}

\resizebox{\linewidth}{!}{
\begin{tabular}{lc|cccccccc} 
\toprule
\multirow{2}{*}{\textbf{Domain}} & \multirow{2}{*}{\textbf{Noise}} & \multicolumn{8}{c}{\textbf{Algorithm}}\\
& & \textbf{PM (ID3)} & \textbf{PM-N (ID3)} & \textbf{PM (C45)} & \textbf{PM-N (C45)} & \textbf{PM (RIPPER)} & \textbf{PM-N (RIPPER)} & \textbf{PM (NSLV)} & \textbf{PM-N (NSLV)}\\
\midrule
\multirow{5}{*}{Blocksworld}
        & 0\%  & \cellcolor{red!25} \xmark & \cellcolor{red!25} \xmark & \cellcolor{red!25} \xmark & \cellcolor{red!25} \xmark & \cellcolor{red!25} \xmark & \cellcolor{red!25} \xmark & \cellcolor{green!25} \cmark & \cellcolor{green!25} \cmark \\
        & 3\%  & \cellcolor{red!25} \xmark & \cellcolor{red!25} \xmark & \cellcolor{red!25} \xmark & \cellcolor{red!25} \xmark & \cellcolor{red!25} \xmark & \cellcolor{red!25} \xmark & \cellcolor{red!25} \xmark & \cellcolor{green!25} \cmark \\
        & 5\% & \cellcolor{red!25} \xmark & \cellcolor{red!25} \xmark & \cellcolor{red!25} \xmark & \cellcolor{red!25} \xmark & \cellcolor{red!25} \xmark & \cellcolor{red!25} \xmark & \cellcolor{red!25} \xmark & \cellcolor{green!25} \cmark \\
        & 10\% & \cellcolor{red!25} \xmark & \cellcolor{red!25} \xmark & \cellcolor{red!25} \xmark & \cellcolor{red!25} \xmark & \cellcolor{red!25} \xmark & \cellcolor{red!25} \xmark & \cellcolor{red!25} \xmark & \cellcolor{green!25} \cmark \\
        & 20\% & \cellcolor{red!25} \xmark & \cellcolor{red!25} \xmark & \cellcolor{red!25} \xmark & \cellcolor{red!25} \xmark & \cellcolor{red!25} \xmark & \cellcolor{red!25} \xmark & \cellcolor{red!25} \xmark & \cellcolor{green!25} \cmark \\
\midrule
\multirow{5}{*}{Depots}
        & 0\% & \cellcolor{red!25} \xmark & \cellcolor{red!25} \xmark & \cellcolor{green!25} \cmark & \cellcolor{green!25} \cmark & \cellcolor{red!25} \xmark & \cellcolor{red!25} \xmark & \cellcolor{green!25} \cmark & \cellcolor{green!25} \cmark \\
        & 3\% & \cellcolor{red!25} \xmark & \cellcolor{red!25} \xmark & \cellcolor{red!25} \xmark & \cellcolor{red!25} \xmark & \cellcolor{red!25} \xmark & \cellcolor{red!25} \xmark & \cellcolor{red!25} \xmark & \cellcolor{green!25} \cmark \\
        & 5\% & \cellcolor{red!25} \xmark & \cellcolor{red!25} \xmark & \cellcolor{red!25} \xmark & \cellcolor{red!25} \xmark & \cellcolor{red!25} \xmark & \cellcolor{red!25} \xmark & \cellcolor{red!25} \xmark & \cellcolor{green!25} \cmark \\
        & 10\% & \cellcolor{red!25} \xmark & \cellcolor{red!25} \xmark & \cellcolor{red!25} \xmark & \cellcolor{red!25} \xmark & \cellcolor{red!25} \xmark & \cellcolor{red!25} \xmark & \cellcolor{red!25} \xmark & \cellcolor{green!25} \cmark \\
        & 20\% & \cellcolor{red!25} \xmark & \cellcolor{red!25} \xmark & \cellcolor{red!25} \xmark & \cellcolor{red!25} \xmark & \cellcolor{red!25} \xmark & \cellcolor{red!25} \xmark & \cellcolor{red!25} \xmark & \cellcolor{green!25} \cmark \\
\midrule
\multirow{5}{*}{DriverLog}
        & 0\% & \cellcolor{red!25} \xmark & \cellcolor{red!25} \xmark & \cellcolor{red!25} \xmark & \cellcolor{red!25} \xmark & \cellcolor{red!25} \xmark & \cellcolor{red!25} \xmark & \cellcolor{green!25} \cmark & \cellcolor{green!25} \cmark \\
        & 3\% & \cellcolor{red!25} \xmark & \cellcolor{red!25} \xmark & \cellcolor{red!25} \xmark & \cellcolor{red!25} \xmark & \cellcolor{red!25} \xmark & \cellcolor{red!25} \xmark & \cellcolor{red!25} \xmark  & \cellcolor{green!25} \cmark \\
        & 5\% & \cellcolor{red!25} \xmark & \cellcolor{red!25} \xmark & \cellcolor{red!25} \xmark & \cellcolor{red!25} \xmark & \cellcolor{red!25} \xmark & \cellcolor{red!25} \xmark & \cellcolor{red!25} \xmark & \cellcolor{green!25} \cmark \\
        & 10\% & \cellcolor{red!25} \xmark & \cellcolor{red!25} \xmark & \cellcolor{red!25} \xmark & \cellcolor{red!25} \xmark & \cellcolor{red!25} \xmark & \cellcolor{red!25} \xmark & \cellcolor{red!25} \xmark & \cellcolor{green!25} \cmark \\
        & 20\% & \cellcolor{red!25} \xmark & \cellcolor{red!25} \xmark & \cellcolor{red!25} \xmark & \cellcolor{red!25} \xmark & \cellcolor{red!25} \xmark & \cellcolor{red!25} \xmark & \cellcolor{red!25} \xmark & \cellcolor{red!25} \xmark \\
\midrule
\multirow{5}{*}{ZenoTravel}
        & 0\% & \cellcolor{red!25} \xmark & \cellcolor{red!25} \xmark & \cellcolor{red!25} \xmark & \cellcolor{red!25} \xmark & \cellcolor{red!25} \xmark & \cellcolor{red!25} \xmark & \cellcolor{green!25} \cmark & \cellcolor{green!25} \cmark \\
        & 3\% & \cellcolor{red!25} \xmark & \cellcolor{red!25} \xmark & \cellcolor{red!25} \xmark & \cellcolor{red!25} \xmark & \cellcolor{red!25} \xmark & \cellcolor{red!25} \xmark & \cellcolor{red!25} \xmark & \cellcolor{green!25} \cmark \\
        & 5\% & \cellcolor{red!25} \xmark  & \cellcolor{red!25} \xmark & \cellcolor{red!25} \xmark & \cellcolor{red!25} \xmark & \cellcolor{red!25} \xmark & \cellcolor{red!25} \xmark & \cellcolor{red!25} \xmark & \cellcolor{green!25} \cmark \\
        & 10\% & \cellcolor{red!25} \xmark & \cellcolor{red!25} \xmark & \cellcolor{red!25} \xmark & \cellcolor{red!25} \xmark & \cellcolor{red!25} \xmark & \cellcolor{red!25} \xmark & \cellcolor{red!25} \xmark & \cellcolor{red!25} \xmark \\
        & 20\% & \cellcolor{red!25} \xmark & \cellcolor{red!25} \xmark & \cellcolor{red!25} \xmark & \cellcolor{red!25} \xmark & \cellcolor{red!25} \xmark & \cellcolor{red!25} \xmark & \cellcolor{red!25} \xmark & \cellcolor{green!25} \cmark \\
\bottomrule
\end{tabular}
}
\caption{Validity Results}
\label{tab:validitySTRIPSC4PM}
\end{table}

In the experiments without noisy elements, those algorithms able to obtain valid planning domains with complete data can learn planning domains with noise-free data. This is because without noise and incompleteness the input data are essentially the same, and PlanMiner-N performs the identically as PlanMiner in the absence of noise in the plan traces. It is when noise is included in the input data that these results begin to diverge, as steep performance drops cause the invalidity of the domains learned by the bulk of the learning algorithms. As a reminder, validity is very sensitive to some domain deficiencies. The lack of a single effect renders the domain totally invalid, and the performance drops of the algorithms are so pronounced that these shortcomings arise everywhere. The exception to this rule is PlanMiner-N (NSLV) which demonstrates its supremacy over all other approaches by learning valid planning domains even in the most complex experiments. PlanMiner-N (NSLV) only fails to obtain valid domains with DriverLog and ZenoTravel in the experiments with the highest percentage of noise. In the case of the DriverLog domain, the invalidity is caused by the creation of a series of spurious preconditions that prevent the correct replication of the test problems, while in the case of ZenoTravel the problem that causes the invalidity is the lack of an effect with a low occurrence rate, which is erroneously detected as noisy by the algorithm and eliminated. 

Finally, in terms of time efficiency, PlanMiner-N is around 10-20\% slower than PlanMiner. This is due to the need to study and apply the noise filter to all predicates, which consumes computational resources.

\subsubsection{State-of-the-art algorithms comparison.} 
\begin{figure}
    \centering
    \resizebox{0.49\linewidth}{!}{
            \begin{tikzpicture}
                \begin{axis}[
                    ylabel={F-Score},
                    xlabel={Noise},
                    ymin=0.2,
                    ymax=1.0,
                    xmin=0,
                    xmax=22,
                    ytick={0.4, 0.5, 0.6, 0.7, 0.8, 0.9, 1.0},
                    xtick={0, 3, 5, 10, 20},
                    legend pos=outer north east,
                    ymajorgrids=true,
                ]
                \addplot[
                    color=red,
                    mark=o,
                    error bars/.cd,
                    y dir=both,
                    y explicit
                    ]
                    coordinates {
                        (0, 0.978)+=(0, 0.007)-=(0, 0.007)
                        (3, 0.764)+=(3, 0.032)-=(3, 0.032)
                        (5, 0.764)+=(5, 0.059)-=(5, 0.059)
                        (10, 0.740)+=(10, 0.061)-=(10, 0.061)
                        (20, 0.640)+=(20, 0.066)-=(20, 0.066)
                    };
                \addlegendentry{ARMS};

                \addplot[
                    color=cyan,
                    mark=o,
                    error bars/.cd,
                    y dir=both,
                    y explicit
                    ]
                    coordinates {
                        (0, 1.0)+=(0, 0.0)-=(0, 0.0)
                        (3,  0.85)+=(3, 0.038)-=(3, 0.038)
                        (5,  0.85)+=(5, 0.038)-=(5, 0.038)
                        (10, 0.75)+=(10, 0.038)-=(10, 0.0389)
                        (20, 0.67)+=(20, 0.039)-=(20, 0.039)
                    };
                \addlegendentry{FAMA};
                
                \addplot[
                    color=olive,
                    mark=o,
                    error bars/.cd,
                    y dir=both,
                    y explicit
                    ]
                    coordinates {
                        (0, 0.91)+=(0, 0.01)-=(0, 0.01)
                        (3,  0.70)+=(3, 0.015)-=(3, 0.015)
                        (5,  0.69)+=(5, 0.017)-=(5, 0.017)
                        (10, 0.64)+=(10, 0.018)-=(10, 0.018)
                        (20, 0.53)+=(20, 0.039)-=(20, 0.039)                        
                    };
                \addlegendentry{Opmaker2};
                
                \addplot[
                    color=orange,
                    mark=o,
                    error bars/.cd,
                    y dir=both,
                    y explicit
                    ]
                    coordinates {
                        (0, 0.96)+=(0, 0.0)-=(0, 0.0)
                        (3,  0.76)+=(3, 0.015)-=(3, 0.015)
                        (5,  0.75)+=(5, 0.017)-=(5, 0.017)
                        (10, 0.74)+=(10, 0.018)-=(10, 0.018)
                        (20, 0.71)+=(20, 0.032)-=(20, 0.032)                        
                    };
                \addlegendentry{AMAN};

                \addplot[
                    color=blue!50,
                    mark=o,
                    error bars/.cd,
                    y dir=both,
                    y explicit
                    ]
                    coordinates {
                        (0, 1)+=(0, 0)-=(0, 0)
                        (3, 1)+=(3, 0)-=(3, 0)
                        (5, 1)+=(5, 0)-=(5, 0)
                        (10, 1)+=(10, 0)-=(10, 0)
                        (20, 1)+=(20, 0)-=(20, 0)
                    };
                \addlegendentry{PM-N (NSLV)};
                
                \end{axis},
            \node[above,font=\large\bfseries] at (current bounding box.north) {BlocksWorld};
            \end{tikzpicture}
    }    
    \hspace{0mm}
    \resizebox{0.49\linewidth}{!}{
            \begin{tikzpicture}
                \begin{axis}[
                    ylabel={F-Score},
                    xlabel={Noise},
                    ymin=0.2,
                    ymax=1.0,
                    xmin=0,
                    xmax=22,
                    ytick={0.4, 0.5, 0.6, 0.7, 0.8, 0.9, 1.0},
                    xtick={0, 3, 5, 10, 20},
                    legend pos=outer north east,
                    ymajorgrids=true,
                ]
                \addplot[
                    color=red,
                    mark=o,
                    error bars/.cd,
                    y dir=both,
                    y explicit
                    ]
                    coordinates {
                        (0, 0.962)+=(0, 0)-=(0, 0)
                        (3, 0.738)+=(3, 0.016)-=(3, 0.016)
                        (5, 0.716)+=(5, 0.016)-=(5, 0.016)
                        (10, 0.579)+=(10, 0.028)-=(10, 0.028)
                        (20, 0.539)+=(20, 0.058)-=(20, 0.058)
                    };
                \addlegendentry{ARMS};

                \addplot[
                    color=cyan,
                    mark=o,
                    error bars/.cd,
                    y dir=both,
                    y explicit
                    ]
                    coordinates {
                        (0, 0.972)+=(0, 0.016)-=(0, 0.016)
                        (3, 0.772)+=(3, 0.025)-=(3, 0.025)
                        (5, 0.762)+=(5, 0.025)-=(5, 0.025)
                        (10, 0.742)+=(10, 0.028)-=(10, 0.028)
                        (20, 0.712)+=(20, 0.078)-=(20, 0.078)
                    };
                \addlegendentry{FAMA};
                
                \addplot[
                    color=olive,
                    mark=o,
                    error bars/.cd,
                    y dir=both,
                    y explicit
                    ]
                    coordinates {
                        (0, 0.93)+=(0, 0.015)-=(0, 0.015)
                        (3, 0.83)+=(3, 0.015)-=(3, 0.015)
                        (5, 0.72)+=(5, 0.017)-=(5, 0.017)
                        (10, 0.62)+=(10, 0.024)-=(10, 0.024)
                        (20, 0.57)+=(20, 0.054)-=(20, 0.054)
                    };
                \addlegendentry{Opmaker2};
                
                \addplot[
                    color=orange,
                    mark=o,
                    error bars/.cd,
                    y dir=both,
                    y explicit
                    ]
                    coordinates {
                        (0, 0.99)+=(0, 0)-=(0, 0)
                        (3, 0.73)+=(3, 0.015)-=(3, 0.015)
                        (5, 0.72)+=(5, 0.017)-=(5, 0.017)
                        (10, 0.70)+=(10, 0.018)-=(10, 0.018)
                        (20, 0.66)+=(20, 0.048)-=(20, 0.048)
                    };
                \addlegendentry{AMAN};
                
                \addplot[
                    color=blue!50,
                    mark=o,
                    error bars/.cd,
                    y dir=both,
                    y explicit
                    ]
                    coordinates {
                        (0, 0.986)+=(0, 0)-=(0, 0)
                        (3, 0.986)+=(3, 0.022)-=(3, 0.022)
                        (5, 0.986)+=(5, 0.050)-=(5, 0.050)
                        (10, 0.985)+=(10, 0.061)-=(10, 0.061)
                        (20, 0.984)+=(20, 0.086)-=(20, 0.086)
                    };
                    
                \addlegendentry{PM-N (NSLV)};
                \end{axis},
            \node[above,font=\large\bfseries] at (current bounding box.north) {Depots};
            \end{tikzpicture}
    }   
    
    \resizebox{0.49\linewidth}{!}{
            \begin{tikzpicture}
                \begin{axis}[
                    ylabel={F-Score},
                    xlabel={Noise},
                    ymin=0.2,
                    ymax=1.0,
                    xmin=0,
                    xmax=22,
                    ytick={0.4, 0.5, 0.6, 0.7, 0.8, 0.9, 1.0},
                    xtick={0, 3, 5, 10, 20},
                    legend pos=outer north east,
                    ymajorgrids=true,
                ]
                
                \addplot[
                    color=red,
                    mark=o,
                    error bars/.cd,
                    y dir=both,
                    y explicit
                    ]
                    coordinates {
                        (0, 0.975)+=(0, 0)-=(0, 0)
                        (3, 0.727)+=(3, 0.007)-=(3, 0.007)
                        (5, 0.681)+=(5, 0.038)-=(5, 0.038)
                        (10, 0.489)+=(10, 0.067)-=(10, 0.067)
                        (20, 0.393)+=(20, 0.095)-=(20, 0.095)
                    };
                \addlegendentry{ARMS};

              \addplot[
                    color=cyan,
                    mark=o,
                    error bars/.cd,
                    y dir=both,
                    y explicit
                    ]
                    coordinates {
                        (0, 0.807)+=(0, 0.016)-=(0, 0.016)
                        (3, 0.592)+=(3, 0.025)-=(3, 0.025)
                        (5, 0.585)+=(5, 0.025)-=(5, 0.025)
                        (10, 0.55)+=(10, 0.028)-=(10, 0.028)
                        (20, 0.448)+=(20, 0.041)-=(20, 0.041)
                    };
                \addlegendentry{FAMA};
                
                \addplot[
                    color=olive,
                    mark=o,
                    error bars/.cd,
                    y dir=both,
                    y explicit
                    ]
                    coordinates {
                        (0, 0.9)+=(0, 0.015)-=(0, 0.015)
                        (3, 0.69)+=(3, 0.015)-=(3, 0.015)
                        (5, 0.67)+=(5, 0.0243)-=(5, 0.0243)
                        (10, 0.65)+=(10, 0.03)-=(10, 0.03)
                        (20, 0.57)+=(20, 0.038)-=(20, 0.038)
                    };
                \addlegendentry{Opmaker2};
                
                \addplot[
                    color=orange,
                    mark=o,
                    error bars/.cd,
                    y dir=both,
                    y explicit
                    ]
                    coordinates {
                        (0, 0.95)+=(0, 0.01)-=(0, 0.01)
                        (3, 0.73)+=(3, 0.015)-=(3, 0.015)
                        (5, 0.65)+=(5, 0.017)-=(5, 0.017)
                        (10, 0.57)+=(10, 0.028)-=(10, 0.028)
                        (20, 0.52)+=(20, 0.039)-=(20, 0.039)
                    };
                \addlegendentry{AMAN};
                
                \addplot[
                    color=blue!50,
                    mark=o,
                    error bars/.cd,
                    y dir=both,
                    y explicit
                    ]
                    coordinates {
                        (0, 0.975)+=(0, 0)-=(0, 0)
                        (3, 0.975)+=(3, 0.000)-=(3, 0.000)
                        (5, 0.974)+=(5, 0.013)-=(5, 0.013)
                        (10, 0.973)+=(10, 0.018)-=(10, 0.018)
                        (20, 0.964)+=(20, 0.053)-=(20, 0.053)
                    };
                    
                \addlegendentry{PM-N (NSLV)};
                \end{axis},
            \node[above,font=\large\bfseries] at (current bounding box.north) {DriverLog};
            \end{tikzpicture}
    }    
    \hspace{0mm}
    \resizebox{0.49\linewidth}{!}{
            \begin{tikzpicture}
                \begin{axis}[
                    ylabel={F-Score},
                    xlabel={Noise},
                    ymin=0.2,
                    ymax=1.0,
                    xmin=0,
                    xmax=22,
                    ytick={0.4, 0.5, 0.6, 0.7, 0.8, 0.9, 1.0},
                    xtick={0, 3, 5, 10, 20},
                    legend pos=outer north east,
                    ymajorgrids=true,
                ]
                
                \addplot[
                    color=red,
                    mark=o,
                    error bars/.cd,
                    y dir=both,
                    y explicit
                    ]
                    coordinates {
                        (0, 0.938)+=(0, 0)-=(0, 0)
                        (3, 0.703)+=(3, 0.015)-=(3, 0.015)
                        (5, 0.691)+=(5, 0.033)-=(5, 0.033)
                        (10, 0.574)+=(10, 0.064)-=(10, 0.064)
                        (20, 0.524)+=(20, 0.098)-=(20, 0.098)
                    };
                \addlegendentry{ARMS};
          
                \addplot[
                    color=cyan,
                    mark=o,
                    error bars/.cd,
                    y dir=both,
                    y explicit
                    ]
                    coordinates {
                        (0, 0.938)+=(0, 0)-=(0, 0)
                        (3, 0.703)+=(3, 0.015)-=(3, 0.015)
                        (5, 0.691)+=(5, 0.033)-=(5, 0.033)
                        (10, 0.594)+=(10, 0.064)-=(10, 0.064)
                        (20, 0.494)+=(20, 0.102)-=(20, 0.102)
                    };
                \addlegendentry{FAMA};
                
                \addplot[
                    color=olive,
                    mark=o,
                    error bars/.cd,
                    y dir=both,
                    y explicit
                    ]
                    coordinates {
                        (0, 0.89)+=(0, 0.015)-=(0, 0.015)
                        (3, 0.67)+=(3, 0.015)-=(3, 0.015)
                        (5, 0.57)+=(5, 0.0243)-=(5, 0.0243)
                        (10, 0.56)+=(10, 0.035)-=(10, 0.035)
                        (20, 0.53)+=(20, 0.065)-=(20, 0.065)
                    };
                \addlegendentry{Opmaker2};
                
                \addplot[
                    color=orange,
                    mark=o,
                    error bars/.cd,
                    y dir=both,
                    y explicit
                    ]
                    coordinates {
                        (0, 0.9)+=(0, 0.01)-=(0, 0.01)
                        (3, 0.87)+=(3, 0.015)-=(3, 0.015)
                        (5, 0.84)+=(5, 0.017)-=(5, 0.017)
                        (10, 0.78)+=(10, 0.028)-=(10, 0.028)
                        (20, 0.69)+=(20, 0.053)-=(20, 0.053)
                    };
                \addlegendentry{AMAN};

                \addplot[
                    color=blue!50,
                    mark=o,
                    error bars/.cd,
                    y dir=both,
                    y explicit
                    ]
                    coordinates {
                        (0, 1.000)+=(0, 0)-=(0, 0)
                        (3, 1.000)+=(3, 0.000)-=(3, 0.000)
                        (5, 1.000)+=(5, 0.000)-=(5, 0.000)
                        (10, 0.984)+=(10, 0.018)-=(10, 0.018)
                        (20, 0.955)+=(20, 0.026)-=(20, 0.026)
                    };
                    
                \addlegendentry{PM-N (NSLV)};
                
                \end{axis},
            \node[above,font=\large\bfseries] at (current bounding box.north) {ZenoTravel};
            \end{tikzpicture}
    }   
    \caption{Performance comparison of between PlanMiner-N and state-of-the-art algorithms on STRIPS domains.}
    \label{fig:compC4STRIPS}
\end{figure}
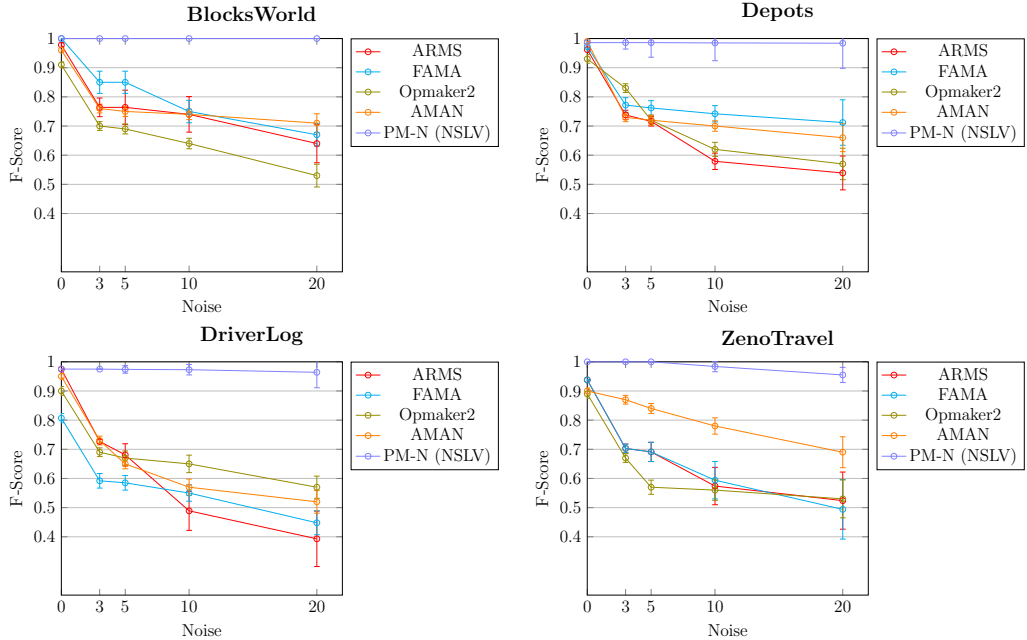

Next, the experimental process is going to study how PlanMiner (NSLV) \textendash the version of PlanMiner with the highest performance\textendash\ performs in comparison to the reference algorithms. Figure \ref{fig:compC4STRIPS} presents a comparative graph that displays the F-Score of these algorithms. Additionally, in Table \ref{tab:validitySTRIPSC4} the validity results of the battery of experiments are shown.  For the sake of readability, to say that the X-axis of the Figure \ref{fig:compC4STRIPS} represents the degree of incompleteness of the input plan traces and that the Tables group the data displayed by the planning domain being learned. For the sake of readibility, the next lines are a summarised version of the experimental result, including only the F-Score and Validity results. 

If we look closely at the results the figure \ref{fig:compC4STRIPS} we can see the following:
\begin{itemize}
    \item \textbf{BlocksWorld.} The benchmark algorithms suffer a steep drop throughout the experimentation, with results 20 to 30 points below the initial values with noise-free data. Of these, OPMaker2 is the worst performer, while ARMS and AMAN maintain similar performance. FAMA, on the other hand, suffers the least from the initial performance drop, but its results worsen severely as the complexity of the experiments increases. PlanMiner-N (NSLV), on the other hand, performs better than all of them, obtaining 100\% F-Score results regardless of the experiment performed.
    \item \textbf{Depots.} Starting from similar initial results, when including noisy data in the input data, the performance of the benchmark algorithms drops by almost 25 points. FAMA and AMAN show some stability in subsequent experiments, while the rest of the state-of-the-art algorithms continue to lose performance down to around 50\% F-Score. In contrast, PlanMiner-N (NSLV) suffers a negligible drop in F-Score throughout the experimentation, maintaining a much higher performance than the benchmark algorithms.
    \item \textbf{DriverLog.} The benchmark algorithms show differences of up to 50 points between their initial results and those obtained in the more complex experiments. As usual, the steepest drops are found in the first experiments, when noise is included in the plan traces. In these experiments, the benchmark algorithms lose 20-25\% F-Score. PlanMiner-N (NSLV) maintains a clear superiority over all benchmark algorithms.
    \item \textbf{ZenoTravel.} The benchmark algorithms present behaviour with large initial drops and constant, but more controlled drops as the experimental conditions are tightened. Of these, AMAN maintains a somewhat more stable behaviour, with more constant drops than the other benchmark algorithms. In the experiments with a higher percentage of noisy elements in the input data, the benchmark algorithms show results close to 50\% F-Score, with the exception of AMAN, whose final results are around 70 points. However, PlanMiner-N (NSLV) shows results almost 30 points higher than AMAN.
\end{itemize}

The benchmark algorithms perform poorly when faced with noise in the input data. This behaviour was to be expected, since, with the exception of AMAN, none of these algorithms was expressly designed to work under noisy situations. Therefore, PlanMiner-N (NSLV) outperforms these algorithms in all experiments. The difference in performance between PlanMiner-N (NSLV) and the reference algorithms ranges from 10-30 points in the experiments with the lowest percentage of noise, to 30-60 points in the experiments with the highest percentage of noise. In terms of performance, second place is held by FAMA and AMAN. When faced with noise, both algorithms suffer a significant loss in performance. AMAN maintains a certain resilience to noise from that point on, showing little variability in results since then, while FAMA does not enjoy this benefit, but its initial performance loss is smaller than AMAN's. This causes FAMA to show slightly better average results. This causes FAMA to show somewhat better average results than AMAN.

\begin{table}
\setlength{\aboverulesep}{-0.5pt}
\setlength{\belowrulesep}{0.0pt}

\resizebox{\linewidth}{!}{
\begin{tabular}{lc|ccccc} 
\toprule
\multirow{2}{*}{\textbf{Domain}} & \multirow{2}{*}{\textbf{Noise}} & \multicolumn{5}{c}{\textbf{Algorithm}}\\
& & \textbf{ARMS} & \textbf{FAMA} & \textbf{OpMaker2} & \textbf{AMAN} & \textbf{PM-N (NSLV)}\\
\midrule
\multirow{5}{*}{Blocksworld}
        & 0\% & \cellcolor{green!25} \cmark & \cellcolor{green!25} \cmark & \cellcolor{green!25} \cmark & \cellcolor{green!25} \cmark & \cellcolor{green!25} \cmark \\
        & 3\% & \cellcolor{red!25} \xmark & \cellcolor{red!25} \xmark & \cellcolor{red!25} \xmark & \cellcolor{red!25} \xmark & \cellcolor{green!25} \cmark \\
        & 5\% & \cellcolor{red!25} \xmark & \cellcolor{red!25} \xmark & \cellcolor{red!25} \xmark & \cellcolor{red!25} \xmark & \cellcolor{green!25} \cmark \\
        & 10\% & \cellcolor{red!25} \xmark & \cellcolor{red!25} \xmark & \cellcolor{red!25} \xmark & \cellcolor{red!25} \xmark & \cellcolor{green!25} \cmark \\
        & 20\% & \cellcolor{red!25} \xmark & \cellcolor{red!25} \xmark & \cellcolor{red!25} \xmark & \cellcolor{red!25} \xmark & \cellcolor{green!25} \cmark \\
\midrule
\multirow{5}{*}{Depots}
        & 0\% & \cellcolor{green!25} \cmark & \cellcolor{green!25} \cmark & \cellcolor{green!25} \cmark & \cellcolor{green!25} \cmark & \cellcolor{green!25} \cmark \\
        & 3\% & \cellcolor{red!25} \xmark & \cellcolor{red!25} \xmark & \cellcolor{red!25} \xmark & \cellcolor{red!25} \xmark & \cellcolor{green!25} \cmark \\
        & 5\% & \cellcolor{red!25} \xmark & \cellcolor{red!25} \xmark & \cellcolor{red!25} \xmark & \cellcolor{red!25} \xmark & \cellcolor{green!25} \cmark \\
        & 10\% & \cellcolor{red!25} \xmark & \cellcolor{red!25} \xmark & \cellcolor{red!25} \xmark & \cellcolor{red!25} \xmark & \cellcolor{green!25} \cmark \\
        & 20\% & \cellcolor{red!25} \xmark & \cellcolor{red!25} \xmark & \cellcolor{red!25} \xmark & \cellcolor{red!25} \xmark & \cellcolor{green!25} \cmark \\
\midrule
\multirow{5}{*}{DriverLog}
        & 0\% & \cellcolor{green!25} \cmark & \cellcolor{red!25} \xmark & \cellcolor{green!25} \cmark & \cellcolor{green!25} \cmark & \cellcolor{green!25} \cmark \\
        & 3\% & \cellcolor{red!25} \xmark & \cellcolor{red!25} \xmark & \cellcolor{red!25} \xmark & \cellcolor{red!25} \xmark & \cellcolor{green!25} \cmark \\
        & 5\% & \cellcolor{red!25} \xmark & \cellcolor{red!25} \xmark & \cellcolor{red!25} \xmark & \cellcolor{red!25} \xmark & \cellcolor{green!25} \cmark \\
        & 10\% & \cellcolor{red!25} \xmark & \cellcolor{red!25} \xmark & \cellcolor{red!25} \xmark & \cellcolor{red!25} \xmark & \cellcolor{green!25} \cmark \\
        & 20\% & \cellcolor{red!25} \xmark & \cellcolor{red!25} \xmark & \cellcolor{red!25} \xmark & \cellcolor{red!25} \xmark & \cellcolor{red!25} \xmark \\
\midrule
\multirow{5}{*}{ZenoTravel}
        & 0\% & \cellcolor{green!25} \cmark & \cellcolor{green!25} \cmark & \cellcolor{red!25} \xmark & \cellcolor{green!25} \cmark & \cellcolor{green!25} \cmark \\
        & 3\% & \cellcolor{red!25} \xmark & \cellcolor{red!25} \xmark & \cellcolor{red!25} \xmark & \cellcolor{red!25} \xmark & \cellcolor{green!25} \cmark \\
        & 5\% & \cellcolor{red!25} \xmark & \cellcolor{red!25} \xmark & \cellcolor{red!25} \xmark & \cellcolor{red!25} \xmark & \cellcolor{green!25} \cmark \\
        & 10\% & \cellcolor{red!25} \xmark & \cellcolor{red!25} \xmark & \cellcolor{red!25} \xmark & \cellcolor{red!25} \xmark & \cellcolor{red!25} \xmark \\
        & 20\% & \cellcolor{red!25} \xmark & \cellcolor{red!25} \xmark & \cellcolor{red!25} \xmark & \cellcolor{red!25} \xmark & \cellcolor{green!25} \cmark \\
\bottomrule
\end{tabular}
}

\caption{Validity Results}
\label{tab:validitySTRIPSC4}
\end{table}

As in the previous experimentation with the different versions of PlanMiner, the generalised loss of performance of the algorithms when noise is included has a negative impact on the validity results of the domains learned with them. Without noise, the benchmark algorithms perform similarly to what we saw in the experimentation presented in the PlanMiner's paper \cite{segura2021planminer}, but when noise is included, this is no longer true, as the multiple errors in the planning domains negatively affect the validity of the domains. 

\subsection{Learning Numerical domains}
\begin{figure}
    \centering
    \resizebox{0.49\linewidth}{!}{
            \begin{tikzpicture}
                \begin{axis}[
                    ylabel={F-Score},
                    xlabel={Noise},
                    ymin=0.2,
                    ymax=1.0,
                    xmin=0,
                    xmax=25,
                    ytick={0.1, 0.2, 0.3, 0.4, 0.5, 0.6, 0.7, 0.8, 0.9, 1.0},
                    xtick={0, 3, 5, 10, 20},
                    legend pos=outer north east,
                    ymajorgrids=true,
                ]
                \addplot[
                    color=magenta,
                    mark=o,
                    error bars/.cd,
                    y dir=both,
                    y explicit
                    ]
                    coordinates {
                        (0, 0.912) +=(0,  0.0)-=(0,  0.0)
                        (3, 0.773) +=(3,  0.057)-=(3,  0.057)
                        (5, 0.653)  +=(5,  0.084)-=(5,  0.084) 
                        (10, 0.489)  +=(10,  0.092)-=(10,  0.092)
                        (20, 0.361)  +=(20,  0.130)-=(20,  0.130)
                    };
                \addlegendentry{PM (C45)};
                \addplot[
                    color=teal,
                    mark=o,
                    error bars/.cd,
                    y dir=both,
                    y explicit
                    ]
                    coordinates {
                        (0, 0.912) +=(0,  0.0)-=(0,  0.0)
                        (3, 0.883) +=(3,  0.050)-=(3,  0.050)
                        (5, 0.745)  +=(5,  0.063)-=(5,  0.063) 
                        (10, 0.699)  +=(10,  0.072)-=(10,  0.072)
                        (20, 0.610)  +=(20,  0.110)-=(20,  0.110)
                    };
                \addlegendentry{PM-N (C45)};
                \addplot[
                    color=blue,
                    mark=o,
                    error bars/.cd,
                    y dir=both,
                    y explicit
                    ]
                    coordinates {
                        (0, 0.865) +=(0,  0.0)-=(0,  0.0)
                        (3, 0.652) +=(3,  0.119)-=(3,  0.119)
                        (5, 0.589)  +=(5,  0.125)-=(5,  0.125) 
                        (10, 0.501)  +=(10,  0.139)-=(10,  0.139)
                        (20, 0.444)  +=(20,  0.176)-=(20,  0.176)
                    };
                \addlegendentry{PM (RIPPER)};
                \addplot[
                    color=cyan,
                    mark=o,
                    error bars/.cd,
                    y dir=both,
                    y explicit
                    ]
                    coordinates {
                        (0, 0.865) +=(0,  0.0)-=(0,  0.0)
                        (3, 0.865) +=(3,  0.00)-=(3,  0.000)
                        (5, 0.865)  +=(5,  0.00)-=(5,  0.00) 
                        (10, 0.812)  +=(10,  0.0386)-=(10,  0.0386)
                        (20, 0.749)  +=(20,  0.117)-=(20,  0.117)
                    };
                \addlegendentry{PM-N (RIPPER)};
                \addplot[
                    color=red,
                    mark=o,
                    error bars/.cd,
                    y dir=both,
                    y explicit
                    ]
                    coordinates {
                        (0, 0.963) +=(0,  0.0)-=(0,  0.0)
                        (3, 0.745) +=(3,  0.046)-=(3,  0.046)
                        (5, 0.736) +=(5,  0.047)-=(5,  0.047)
                        (10, 0.706) +=(10,  0.049)-=(10,  0.049)         
        	            (20, 0.690) +=(20,  0.049)-=(20,  0.049)
                    };
                \addlegendentry{PM (NSLV)};
                \addplot[
                    color=olive,
                    mark=o,
                    error bars/.cd,
                    y dir=both,
                    y explicit
                    ]
                    coordinates {
                        (0, 0.963) +=(0,  0.0)-=(0,  0.0)
                        (3, 0.946) +=(3,  0.070)-=(3,  0.070)
                        (5, 0.929)  +=(5,  0.091)-=(5,  0.091) 
                        (10, 0.882)  +=(10,  0.091)-=(10,  0.091)
                        (20, 0.866)  +=(20,  0.100)-=(20,  0.100)
                    };
                \addlegendentry{PM-N (NSLV)};
                \end{axis},
            \node[above,font=\large\bfseries] at (current bounding box.north) {Depots};
            \end{tikzpicture}
    }    
    \hspace{0mm}
    \resizebox{0.49\linewidth}{!}{
            \begin{tikzpicture}
                \begin{axis}[
                    ylabel={F-Score},
                    xlabel={Noise},
                    ymin=0.2,
                    ymax=1.0,
                    xmin=0,
                    xmax=25,
                    ytick={0.1, 0.2, 0.3, 0.4, 0.5, 0.6, 0.7, 0.8, 0.9, 1.0},
                    xtick={0, 3, 5, 10, 20},
                    legend pos=outer north east,
                    ymajorgrids=true,
                ]
                \addplot[
                    color=magenta,
                    mark=o,
                    error bars/.cd,
                    y dir=both,
                    y explicit
                    ]
                    coordinates {
                        (0, 0.932) +=(0,  0.0)-=(0,  0.0)
                        (3, 0.7006) +=(3,  0.063)-=(3,  0.063)
                        (5, 0.652)  +=(5,  0.0957)-=(5,  0.0957) 
                        (10, 0.620)  +=(10,  0.109)-=(10,  0.109)
                        (20, 0.537)  +=(20,  0.136)-=(20,  0.136)
                    };
                \addlegendentry{PM (C45)};
                \addplot[
                    color=teal,
                    mark=o,
                    error bars/.cd,
                    y dir=both,
                    y explicit
                    ]
                    coordinates {
                        (0, 0.932) +=(0,  0.0)-=(0,  0.0)
                        (3, 0.902) +=(3,  0.006)-=(3,  0.006)
                        (5, 0.898)  +=(5,  0.025)-=(5,  0.025) 
                        (10, 0.862)  +=(10,  0.048)-=(10,  0.048)
                        (20, 0.795)  +=(20,  0.069)-=(20,  0.069)
                    };
                \addlegendentry{PM-N (C45)};
                \addplot[
                    color=blue,
                    mark=o,
                    error bars/.cd,
                    y dir=both,
                    y explicit
                    ]
                    coordinates {
                        (0, 0.9001) +=(0,  0.0)-=(0,  0.0)
                        (3, 0.682) +=(3,  0.152)-=(3,  0.152)
                        (5, 0.587)  +=(5,  0.167)-=(5,  0.167) 
                        (10, 0.438)  +=(10,  0.179)-=(10,  0.179)
                        (20, 0.384)  +=(20,  0.183)-=(20,  0.183)
                    };
                \addlegendentry{PM (RIPPER)};
                \addplot[
                    color=cyan,
                    mark=o,
                    error bars/.cd,
                    y dir=both,
                    y explicit
                    ]
                    coordinates {
                        (0, 0.9001) +=(0,  0.0)-=(0,  0.0)
                        (3, 0.860) +=(3,  0.071)-=(3,  0.071)
                        (5, 0.859)  +=(5,  0.073)-=(5,  0.073) 
                        (10, 0.813)  +=(10,  0.086)-=(10,  0.086)
                        (20, 0.769)  +=(20,  0.098)-=(20,  0.098)
                    };
                \addlegendentry{PM-N (RIPPER)};
                \addplot[
                    color=red,
                    mark=o,
                    error bars/.cd,
                    y dir=both,
                    y explicit
                    ]
                    coordinates {
                        (0, 0.948) +=(0,  0.0)-=(0,  0.0)
                        (3, 0.678) +=(3,  0.045)-=(3,  0.045)
                        (5, 0.665) +=(5,  0.048)-=(5,  0.048)
                        (10, 0.637) +=(10,  0.069)-=(10,  0.069)         
        	            (20, 0.553) +=(20,  0.091)-=(20,  0.091)
                    };
                \addlegendentry{PM (NSLV)};
                \addplot[
                    color=olive,
                    mark=o,
                    error bars/.cd,
                    y dir=both,
                    y explicit
                    ]
                    coordinates {
                        (0, 0.948) +=(0,  0.0)-=(0,  0.0)
                        (3, 0.946) +=(3,  0.018)-=(3,  0.018)
                        (5, 0.926)  +=(5,  0.041)-=(5,  0.041) 
                        (10, 0.903)  +=(10,  0.047)-=(10,  0.047)
                        (20, 0.863)  +=(20,  0.072)-=(20,  0.072)
                    };
                \addlegendentry{PM-N (NSLV)};
                \end{axis},
            \node[above,font=\large\bfseries] at (current bounding box.north) {DriverLog};
            \end{tikzpicture}
    }
    \resizebox{0.49\linewidth}{!}{
            \begin{tikzpicture}
                \begin{axis}[
                    ylabel={F-Score},
                    xlabel={Noise},
                    ymin=0.2,
                    ymax=1.0,
                    xmin=0,
                    xmax=25,
                    ytick={0.1, 0.2, 0.3, 0.4, 0.5, 0.6, 0.7, 0.8, 0.9, 1.0},
                    xtick={0, 3, 5, 10, 20},
                    legend pos=outer north east,
                    ymajorgrids=true,
                ]
                \addplot[
                    color=magenta,
                    mark=o,
                    error bars/.cd,
                    y dir=both,
                    y explicit
                    ]
                    coordinates {
                        (0, 0.839) +=(0,  0.0)-=(0,  0.0)
                        (3, 0.799) +=(3,  0.079)-=(3,  0.079)
                        (5, 0.783)  +=(5,  0.082)-=(5,  0.082) 
                        (10, 0.676)  +=(10,  0.100)-=(10,  0.100)
                        (20, 0.527)  +=(20,  0.134)-=(20,  0.134)
                    };
                \addlegendentry{PM (C45)};
                \addplot[
                    color=teal,
                    mark=o,
                    error bars/.cd,
                    y dir=both,
                    y explicit
                    ]
                    coordinates {
                        (0, 0.839) +=(0,  0.0)-=(0,  0.0)
                        (3, 0.811) +=(3,  0.052)-=(3,  0.052)
                        (5, 0.803)  +=(5,  0.058)-=(5,  0.058) 
                        (10, 0.742)  +=(10,  0.077)-=(10,  0.077)
                        (20, 0.718)  +=(20,  0.082)-=(20,  0.082)
                    };
                \addlegendentry{PM-N (C45)};
                \addplot[
                    color=blue,
                    mark=o,
                    error bars/.cd,
                    y dir=both,
                    y explicit
                    ]
                    coordinates {
                        (0, 0.844) +=(0,  0.0)-=(0,  0.0)
                        (3, 0.705) +=(3,  0.0554)-=(3,  0.0554)
                        (5, 0.630)  +=(5,  0.0872)-=(5,  0.0872) 
                        (10, 0.577)  +=(10,  0.116)-=(10,  0.116)
                        (20, 0.499)  +=(20,  0.138)-=(20,  0.138)
                    };
                \addlegendentry{PM (RIPPER)};
                \addplot[
                    color=cyan,
                    mark=o,
                    error bars/.cd,
                    y dir=both,
                    y explicit
                    ]
                    coordinates {
                        (0, 0.844) +=(0,  0.0)-=(0,  0.0)
                        (3, 0.829) +=(3,  0.024)-=(3,  0.024)
                        (5, 0.815)  +=(5,  0.035)-=(5,  0.035) 
                        (10, 0.773)  +=(10,  0.046)-=(10,  0.046)
                        (20, 0.739)  +=(20,  0.068)-=(20,  0.068)
                    };
                \addlegendentry{PM-N (RIPPER)};
                \addplot[
                    color=red,
                    mark=o,
                    error bars/.cd,
                    y dir=both,
                    y explicit
                    ]
                    coordinates {
                        (0, 0.873) +=(0,  0.0)-=(0,  0.0)
                        (3, 0.463) +=(3,  0.033)-=(3,  0.033)
                        (5, 0.417) +=(5,  0.040)-=(5,  0.040)
                        (10, 0.402) +=(10,  0.042)-=(10,  0.042)         
        	            (20, 0.379) +=(20,  0.064)-=(20,  0.064)
                    };
                \addlegendentry{PM (NSLV)};
                \addplot[
                    color=olive,
                    mark=o,
                    error bars/.cd,
                    y dir=both,
                    y explicit
                    ]
                    coordinates {
                        (0, 0.873) +=(0,  0.0)-=(0,  0.0)
                        (3, 0.863) +=(3,  0.033)-=(3,  0.033)
                        (5, 0.817)  +=(5,  0.040)-=(5,  0.040) 
                        (10, 0.802)  +=(10,  0.042)-=(10,  0.042)
                        (20, 0.779)  +=(20,  0.064)-=(20,  0.064)
                    };
                \addlegendentry{PM-N (NSLV)};
                \end{axis},
            \node[above,font=\large\bfseries] at (current bounding box.north) {Rovers};
            \end{tikzpicture}
    }
    \hspace{0mm}
    \resizebox{0.49\linewidth}{!}{
            \begin{tikzpicture}
                \begin{axis}[
                    ylabel={F-Score},
                    xlabel={Noise},
                    ymin=0.2,
                    ymax=1.0,
                    xmin=0,
                    xmax=25,
                    ytick={0.1, 0.2, 0.3, 0.4, 0.5, 0.6, 0.7, 0.8, 0.9, 1.0},
                    xtick={0, 3, 5, 10, 20},
                    legend pos=outer north east,
                    ymajorgrids=true,
                ]
                \addplot[
                    color=magenta,
                    mark=o,
                    error bars/.cd,
                    y dir=both,
                    y explicit
                    ]
                    coordinates {
                        (0, 0.974) +=(0,  0.0)-=(0,  0.0)
                        (3, 0.724) +=(3,  0.0477)-=(3,  0.0477)
                        (5, 0.695)  +=(5,  0.0738)-=(5,  0.0738) 
                        (10, 0.646)  +=(10,  0.100)-=(10,  0.100)
                        (20, 0.572)  +=(20,  0.117)-=(20,  0.117)
                    };
                \addlegendentry{PM (C45)};
                \addplot[
                    color=teal,
                    mark=o,
                    error bars/.cd,
                    y dir=both,
                    y explicit
                    ]
                    coordinates {
                        (0, 0.974) +=(0,  0.0)-=(0,  0.0)
                        (3, 0.925) +=(3,  0.026)-=(3,  0.026)
                        (5, 0.903)  +=(5,  0.031)-=(5,  0.031) 
                        (10, 0.868)  +=(10,  0.044)-=(10,  0.044)
                        (20, 0.825)  +=(20,  0.069)-=(20,  0.069)
                    };
                \addlegendentry{PM-N (C45)};
                \addplot[
                    color=blue,
                    mark=o,
                    error bars/.cd,
                    y dir=both,
                    y explicit
                    ]
                    coordinates {
                        (0, 0.942) +=(0,  0.0)-=(0,  0.0)
                        (3, 0.703) +=(3,  0.0627)-=(3,  0.0627)
                        (5, 0.671)  +=(5,  0.0994)-=(5,  0.0994) 
                        (10, 0.542)  +=(10,  0.125)-=(10,  0.0125)
                        (20, 0.488)  +=(20,  0.148)-=(20,  0.148)
                    };
                \addlegendentry{PM (RIPPER)};
                \addplot[
                    color=cyan,
                    mark=o,
                    error bars/.cd,
                    y dir=both,
                    y explicit
                    ]
                    coordinates {
                        (0, 0.942) +=(0,  0.0)-=(0,  0.0)
                        (3, 0.914) +=(3,  0.026)-=(3,  0.026)
                        (5, 0.889)  +=(5,  0.031)-=(5,  0.031) 
                        (10, 0.852)  +=(10,  0.033)-=(10,  0.033)
                        (20, 0.800)  +=(20,  0.046)-=(20,  0.046)
                    };
                \addlegendentry{PM-N (RIPPER)};
                \addplot[
                    color=red,
                    mark=o,
                    error bars/.cd,
                    y dir=both,
                    y explicit
                    ]
                    coordinates {
                        (0, 1.000) +=(0,  0.0)-=(0,  0.0)
                        (3, 0.569) +=(3,  0.056)-=(3,  0.056)
                        (5, 0.490) +=(5,  0.059)-=(5,  0.059)
                        (10, 0.487) +=(10,  0.069)-=(10,  0.069)         
        	            (20, 0.460) +=(20,  0.077)-=(20,  0.077)
                    };
                \addlegendentry{PM (NSLV)};
                \addplot[
                    color=olive,
                    mark=o,
                    error bars/.cd,
                    y dir=both,
                    y explicit
                    ]
                    coordinates {
                        (0, 1.000) +=(0,  0.0)-=(0,  0.0)
                        (3, 0.972) +=(3,  0.015)-=(3,  0.015)
                        (5, 0.972)  +=(5,  0.015)-=(5,  0.015) 
                        (10, 0.952)  +=(10,  0.019)-=(10,  0.019)
                        (20, 0.883)  +=(20,  0.029)-=(20,  0.029)
                    };
                \addlegendentry{PM-N (NSLV)};
                \end{axis},
            \node[above,font=\large\bfseries] at (current bounding box.north) {Satellite};
            \end{tikzpicture}
    }
    \resizebox{0.49\linewidth}{!}{
            \begin{tikzpicture}
                \begin{axis}[
                    ylabel={F-Score},
                    xlabel={Noise},
                    ymin=0.2,
                    ymax=1.0,
                    xmin=0,
                    xmax=25,
                    ytick={0.1, 0.2, 0.3, 0.4, 0.5, 0.6, 0.7, 0.8, 0.9, 1.0},
                    xtick={0, 3, 5, 10, 20},
                    legend pos=outer north east,
                    ymajorgrids=true,
                ]
                \addplot[
                    color=magenta,
                    mark=o,
                    error bars/.cd,
                    y dir=both,
                    y explicit
                    ]
                    coordinates {
                        (0, 0.868) +=(0,  0.0)-=(0,  0.0)
                        (3, 0.602) +=(3,  0.0927)-=(3,  0.0927)
                        (5, 0.538)  +=(5,  0.126)-=(5,  0.126) 
                        (10, 0.465)  +=(10,  0.131)-=(10,  0.131)
                        (20, 0.399)  +=(20,  0.147)-=(20,  0.147)
                    };
                \addlegendentry{PM (C45)};
                \addplot[
                    color=teal,
                    mark=o,
                    error bars/.cd,
                    y dir=both,
                    y explicit
                    ]
                    coordinates {
                        (0, 0.868) +=(0,  0.0)-=(0,  0.0)
                        (3, 0.842) +=(3,  0.027)-=(3,  0.027)
                        (5, 0.822)  +=(5,  0.036)-=(5,  0.036) 
                        (10, 0.818)  +=(10,  0.059)-=(10,  0.059)
                        (20, 0.767)  +=(20,  0.073)-=(20,  0.073)
                    };
                \addlegendentry{PM-N (C45)};
                \addplot[
                    color=blue,
                    mark=o,
                    error bars/.cd,
                    y dir=both,
                    y explicit
                    ]
                    coordinates {
                        (0, 0.855) +=(0,  0.0)-=(0,  0.0)
                        (3, 0.424) +=(3,  0.164)-=(3,  0.164)
                        (5, 0.403)  +=(5,  0.166)-=(5,  0.166) 
                        (10, 0.396)  +=(10,  0.169)-=(10,  0.169)
                        (20, 0.300)  +=(20,  0.183)-=(20,  0.183)
                    };
                \addlegendentry{PM (RIPPER)};
                \addplot[
                    color=cyan,
                    mark=o,
                    error bars/.cd,
                    y dir=both,
                    y explicit
                    ]
                    coordinates {
                        (0, 0.855) +=(0,  0.0)-=(0,  0.0)
                        (3, 0.826) +=(3,  0.024)-=(3,  0.024)
                        (5, 0.796)  +=(5,  0.031)-=(5,  0.031) 
                        (10, 0.761)  +=(10,  0.038)-=(10,  0.038)
                        (20, 0.6995)  +=(20,  0.045)-=(20,  0.045)
                    };
                \addlegendentry{PM-N (RIPPER)};
                \addplot[
                    color=red,
                    mark=o,
                    error bars/.cd,
                    y dir=both,
                    y explicit
                    ]
                    coordinates {
                        (0, 0.896) +=(0,  0.0)-=(0,  0.0)
                        (3, 0.546) +=(3,  0.045)-=(3,  0.045)
                        (5, 0.510) +=(5,  0.056)-=(5,  0.056)
                        (10, 0.477) +=(10,  0.061)-=(10,  0.061)         
        	            (20, 0.458) +=(20,  0.065)-=(20,  0.065)
                    };
                \addlegendentry{PM (NSLV)};
                \addplot[
                    color=olive,
                    mark=o,
                    error bars/.cd,
                    y dir=both,
                    y explicit
                    ]
                    coordinates {
                        (0, 0.896) +=(0,  0.0)-=(0,  0.0)
                        (3, 0.892) +=(3,  0.016)-=(3,  0.016)
                        (5, 0.848)  +=(5,  0.028)-=(5,  0.028) 
                        (10, 0.819)  +=(10,  0.042)-=(10,  0.042)
                        (20, 0.781)  +=(20,  0.059)-=(20,  0.059)
                    };
                \addlegendentry{PM-N (NSLV)};
                \end{axis},
            \node[above,font=\large\bfseries] at (current bounding box.north) {ZenoTravel};
            \end{tikzpicture}
    }
    \caption{Performance comparison of PlanMiner-N using different classification algorithms on Numerical domains.}
    \label{fig:compC4NUM}
\end{figure}
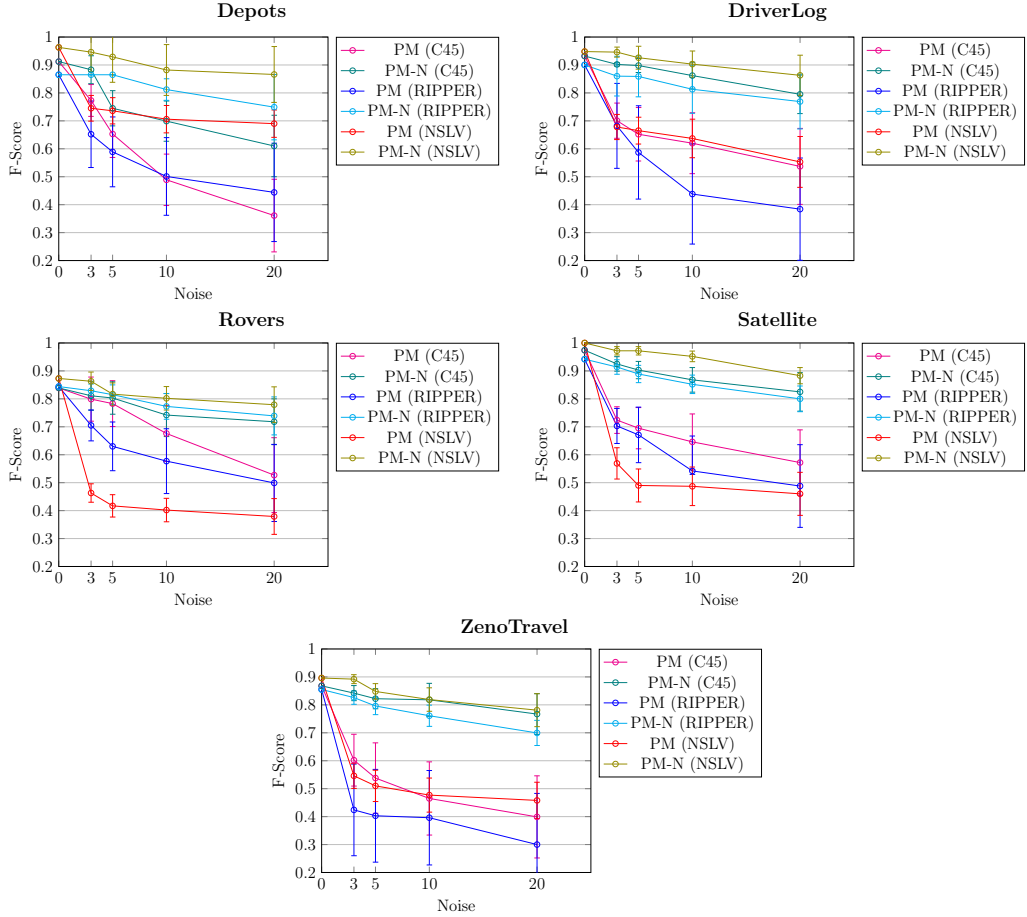

Finally, the experimental process studies how the selected classification algorithm affects the performance of PlanMiner-N when learning planning domains with numerical information. The classification algorithms used in this experimentation are the same used in the last experimental process, except for ID3. The results of this experimentation are divided between the Figure \ref{fig:compC4NUM} and the Table \ref{tab:validityNUMC4}. First, Figure \ref{fig:compC4NUM} contains a graph that compares the algorithms in terms of F-Score, while, Table \ref{tab:validityNUMC4} contains the validity results of PlanMiner with the different classification algorithms. In order to improve the readability of the results, Figure \ref{fig:compC4NUM} set the incompleteness degree of the plan traces in the X-axis, and Table \ref{tab:validityNUMC4} shorts the results given the domain from which they were obtained. As with the experimental blocks presented above, this section presents a summary version of the results. 

If we look closely at the results in Figure \ref{fig:compC4NUM} we can see the following:
\begin{itemize}
    \item \textbf{Depots.} PlanMiner-N (NSLV) maintains 88\% F-Score levels even in the complex experiments, losing less than 10\% F-Score compared to the noiseless results. PlanMiner (NSLV) on the other hand suffers a drop of 20 using 3\% of noisy elements, its final performance is below 70\% F-Score in the experiments with 20\% of noisy information. PlanMiner-N (C45) suffers a large F-Score loss of 13 points in the experiment with 5\% noisy data, which stabilises somewhat in the following experiments. PlanMiner (C45), on the other hand, suffers a steady drop in performance, bringing its final F-Score below 40 points. PlanMiner-N (RIPPER) and PlanMiner (RIPPER) show the biggest difference of all the algorithms, since, where PlanMiner-N (RIPPER) remains constant throughout the experimentation, suffering only F-Score losses in the more complex experiments, PlanMiner (RIPPER) drops steadily.
    \item \textbf{DriverLog.} The results of PlanMiner-N (C45), PlanMiner-N (RIPPER) and PlanMiner-N (NSLV) is similar regardless of the classification algorithm used. All approaches show a slight drop in F-Score throughout the experimentation, with intervals where the F-Score loss is zero. In general, these approximations lose on average 6 F-Score points between the experimentation without noisy data and the one with more noisy data.  On the other hand, the experimentation on PlanMiner (C45), PlanMiner (RIPPER) and PlanMiner (NSLV) shows a steep F-Score drop of more than 20 points when noise is included. PlanMiner (C45) and PlanMiner (NSLV) moderate this tendency a little to reach an F-Score of around 55\% in the final experiments, PlanMiner (RIPPER) on the other hand does not, reaching values below 40 F-Score points.
    \item \textbf{Rovers.} PlanMiner-N (NSLV) is little affected by noise, suffering a significant F-Score loss of 5 points in the experimentation with 5\% noisy data. In the rest of the experimentation, the F-Score loss is minimal. PlanMiner (NSLV) loses 40\% F-Score when noise is found in the input traces, a big contrast when compared to the results of PlanMiner-N (NSLV). Although the other proposals behave similarly, the F-Score loss of PlanMiner (NSLV) is the most pronounced in the whole experiment. PlanMiner-N (C45) performs at 72 points, compared to PlanMiner (C45) at 51 points, and PlanMiner-N (RIPPER) with a final F-Score of 73\%, 20 points higher than PlanMiner (RIPPER).
    \item \textbf{Satellite.} Similar to DriverLog, the results of PlanMiner-N (C45), PlanMiner-N (RIPPER) and PlanMiner-N (NSLV) show a similar trend: a slight but steady drop in F-Score as the experimental assumptions become more complicated. These algorithms show F-Score drops of 11\% in the more complex experiments compared to the no noise experiment. The results of the PlanMiner experiment show drops of 25 points with the C45 classifier and RIPPER, and more than 40 points with NSLV when faced with some noise. PlanMiner (NSLV) maintains the results in the rest of the experimental assumptions, leaving its final performance at 48 points, while PlanMiner (C45) and PlanMiner (RIPPER) obtain an F-Score of 58\% and 49\% respectively.  
    \item \textbf{ZenoTravel.} PlanMiner-N (NSLV) remains unchanged at certain noise levels, with an F-Score of 89\% at 3\% noise. PlanMiner-N (NSLV) loses around 11\% F-Score in the more complex experimentation. PlanMiner (RIPPER) and (C45) lose some F-Score until experimentation with 5\% noise, but while PlanMiner (C45) shows some resistance to noise, (RIPPER) loses performance steadily. On the other hand, PlanMiner (NSLV), PlanMiner (C45) and PlanMiner (RIPPER) show a large initial drop, followed by a certain stabilisation of the results. The F-Score value in the most complex experiments is 30 points PlanMiner (RIPPER), 40 points PlanMiner (C45) and 47 points PlanMiner (NSLV).
\end{itemize}

PlanMiner demonstrates some ``natural resistance'' to noise in this experiments, showing somewhat better results than the experiments with STRIPS domains. Still, the difference in performance between the version of PlanMiner presented in this paper and the previous version is remarkable. Using the C.45 classifier, PlanMiner-N performs around 30 points better than PlanMiner, which can reach up to 50 points difference in certain planning domains. On the other hand, with the RIPPER classification algorithm, it maintains in all experiments a higher performance than 70\% with PlanMiner-N, while with PlanMiner it reaches values around 30 F-Score points. Finally, with NSLV the same process is repeated (i.e. F-Score differences close to 4\%), although aggravated with very significant decreases in performance simply by including noise. This contrasts sharply with PlanMiner-N (NSLV) which remains relatively unperturbed by noise, far outperforming the other approaches tested in the experiment. The aforementioned ``natural resistance'' to noise is given by the way the feature discovery component works, namely it is a product of the design of the symbolic regressor. Since the symbolic regressor does not seek exact results, but rather takes approximate (but as close to exact as possible) results as correct, the effect of the inclusion of certain noisy elements is diluted. Although these elements increase the error of the expression that is being learned, it is possible that the expression meets the acceptance criteria of the algorithm and is accepted by the learning algorithm. Unfortunately, this method is not infallible, and if insufficient data is available or if there are outliers with the potential to greatly perturb the error calculation, PlanMiner is unable to correctly learn the target expression. The inclusion of noise treatment methods increases this resistance, not so much because they influence the behaviour of the regression algorithm, but because they alter the noise problem and shift it to an incompleteness problem. As seen in the empirical studies of PlanMiner \cite{segura2021planminer}, incompleteness is highly tolerated by PlanMiner (and thus by PlanMiner-N) improving greatly the performance of the algorithm. Looking at the evolution of the accuracy and recall metrics of the algorithms throughout the experimentation, we can see how noise affects the first metric much more than the second. This is because any amount of noisy elements not addressed by the noise filtering processes triggers the bias problems described in the paper of PlanMiner. As we have seen in previous experiments, the property of NSLV to generate descriptive rules plays in its favour against bias problems. This particular case is no exception, and it is the main reason why NSLV performs better than the other classifiers. 

\begin{table}
\setlength{\aboverulesep}{-0.5pt}
\setlength{\belowrulesep}{0.0pt}

\resizebox{\linewidth}{!}{
\begin{tabular}{lc|cccccc} 
\toprule
\multirow{2}{*}{\textbf{Domain}} & \multirow{2}{*}{\textbf{Noise}} & \multicolumn{6}{c}{\textbf{Algorithm}}\\
& & \textbf{PM (C4.5)} & \textbf{PM-N (C4.5)} & \textbf{PM (RIPPER)} & \textbf{PM-N (RIPPER)} & \textbf{PM (NSLV)} & \textbf{PM-N (NSLV)} \\

\midrule
\multirow{5}{*}{Depots}
        & 0\% & \cellcolor{green!25} \cmark & \cellcolor{green!25} \cmark & \cellcolor{red!25} \xmark & \cellcolor{red!25} \xmark & \cellcolor{green!25} \cmark & \cellcolor{green!25} \cmark \\
        & 3\% & \cellcolor{red!25} \xmark & \cellcolor{green!25} \cmark & \cellcolor{red!25} \xmark & \cellcolor{red!25} \xmark & \cellcolor{red!25} \xmark & \cellcolor{green!25} \cmark \\
        & 5\% & \cellcolor{red!25} \xmark & \cellcolor{red!25} \xmark & \cellcolor{red!25} \xmark & \cellcolor{red!25} \xmark & \cellcolor{red!25} \xmark & \cellcolor{red!25} \xmark \\
        & 10\% & \cellcolor{red!25} \xmark & \cellcolor{red!25} \xmark & \cellcolor{red!25} \xmark & \cellcolor{red!25} \xmark & \cellcolor{red!25} \xmark & \cellcolor{red!25} \xmark \\
        & 20\% & \cellcolor{red!25} \xmark & \cellcolor{red!25} \xmark & \cellcolor{red!25} \xmark & \cellcolor{red!25} \xmark & \cellcolor{red!25} \xmark & \cellcolor{red!25} \xmark \\
\midrule
\multirow{5}{*}{DriverLog}
        & 0\% & \cellcolor{red!25} \xmark & \cellcolor{red!25} \xmark & \cellcolor{red!25} \xmark & \cellcolor{red!25} \xmark & \cellcolor{green!25} \cmark & \cellcolor{green!25} \cmark \\
        & 3\% & \cellcolor{red!25} \xmark & \cellcolor{red!25} \xmark & \cellcolor{red!25} \xmark & \cellcolor{red!25} \xmark & \cellcolor{red!25} \xmark & \cellcolor{green!25} \cmark \\
        & 5\% & \cellcolor{red!25} \xmark & \cellcolor{red!25} \xmark & \cellcolor{red!25} \xmark & \cellcolor{red!25} \xmark & \cellcolor{red!25} \xmark & \cellcolor{red!25} \xmark \\
        & 10\% & \cellcolor{red!25} \xmark & \cellcolor{red!25} \xmark & \cellcolor{red!25} \xmark & \cellcolor{red!25} \xmark & \cellcolor{red!25} \xmark & \cellcolor{red!25} \xmark \\
        & 20\% & \cellcolor{red!25} \xmark & \cellcolor{red!25} \xmark & \cellcolor{red!25} \xmark & \cellcolor{red!25} \xmark & \cellcolor{red!25} \xmark & \cellcolor{red!25} \xmark \\
\midrule
\multirow{5}{*}{Rovers}
        & 0\% & \cellcolor{red!25} \xmark & \cellcolor{red!25} \xmark & \cellcolor{green!25} \cmark & \cellcolor{green!25} \cmark & \cellcolor{green!25} \cmark & \cellcolor{green!25} \cmark \\
        & 3\% & \cellcolor{red!25} \xmark & \cellcolor{red!25} \xmark & \cellcolor{red!25} \xmark & \cellcolor{red!25} \xmark & \cellcolor{red!25} \xmark & \cellcolor{green!25} \cmark \\
        & 5\% & \cellcolor{red!25} \xmark & \cellcolor{red!25} \xmark & \cellcolor{red!25} \xmark & \cellcolor{red!25} \xmark & \cellcolor{red!25} \xmark & \cellcolor{red!25} \xmark \\
        & 10\% & \cellcolor{red!25} \xmark & \cellcolor{red!25} \xmark & \cellcolor{red!25} \xmark & \cellcolor{red!25} \xmark & \cellcolor{red!25} \xmark & \cellcolor{red!25} \xmark \\
        & 20\% & \cellcolor{red!25} \xmark & \cellcolor{red!25} \xmark & \cellcolor{red!25} \xmark & \cellcolor{red!25} \xmark & \cellcolor{red!25} \xmark & \cellcolor{red!25} \xmark \\
\midrule
\multirow{5}{*}{Satellite}
       & 0\% & \cellcolor{red!25} \xmark & \cellcolor{red!25} \xmark & \cellcolor{green!25} \cmark & \cellcolor{green!25} \cmark & \cellcolor{green!25} \cmark & \cellcolor{green!25} \cmark \\
        & 3\% & \cellcolor{red!25} \xmark & \cellcolor{red!25} \xmark & \cellcolor{green!25} \cmark & \cellcolor{green!25} \cmark & \cellcolor{red!25} \xmark & \cellcolor{green!25} \cmark \\
        & 5\% & \cellcolor{red!25} \xmark & \cellcolor{red!25} \xmark & \cellcolor{red!25} \xmark & \cellcolor{green!25} \cmark & \cellcolor{red!25} \xmark & \cellcolor{green!25} \cmark \\
        & 10\% & \cellcolor{red!25} \xmark & \cellcolor{red!25} \xmark & \cellcolor{red!25} \xmark & \cellcolor{red!25} \xmark & \cellcolor{red!25} \xmark & \cellcolor{green!25} \cmark \\
        & 20\% & \cellcolor{red!25} \xmark & \cellcolor{red!25} \xmark & \cellcolor{red!25} \xmark & \cellcolor{red!25} \xmark & \cellcolor{red!25} \xmark & \cellcolor{red!25} \xmark \\
\midrule
\multirow{5}{*}{ZenoTravel}
       & 0\% & \cellcolor{red!25} \xmark & \cellcolor{red!25} \xmark & \cellcolor{red!25} \xmark & \cellcolor{red!25} \xmark & \cellcolor{green!25} \cmark & \cellcolor{green!25} \cmark \\
        & 3\% & \cellcolor{red!25} \xmark & \cellcolor{red!25} \xmark & \cellcolor{red!25} \xmark & \cellcolor{red!25} \xmark & \cellcolor{red!25} \xmark & \cellcolor{green!25} \cmark \\
        & 5\% & \cellcolor{red!25} \xmark & \cellcolor{red!25} \xmark & \cellcolor{red!25} \xmark & \cellcolor{red!25} \xmark & \cellcolor{red!25} \xmark & \cellcolor{red!25} \xmark \\
        & 10\% & \cellcolor{red!25} \xmark & \cellcolor{red!25} \xmark & \cellcolor{red!25} \xmark & \cellcolor{red!25} \xmark & \cellcolor{red!25} \xmark & \cellcolor{red!25} \xmark \\
        & 20\% & \cellcolor{red!25} \xmark & \cellcolor{red!25} \xmark & \cellcolor{red!25} \xmark & \cellcolor{red!25} \xmark & \cellcolor{red!25} \xmark & \cellcolor{red!25} \xmark \\
\bottomrule
\end{tabular}
}

\caption{Validity Results}
\label{tab:validityNUMC4}
\end{table}

Validity experiments show again the effectiveness of the methods implemented in PlanMiner-N for dealing with noise. If we compare the results of PlanMiner and PlanMiner-N, we see that PlanMiner's ``natural resistance'' to noise when learning numerical domains is not infallible and does not guarantee obtaining valid domains. The reason for this is as previously indicated earlier in the experimentation: the validity criteria are very demanding, and a single erroneous effect causes the entire planning domain to be invalid. PlanMiner can meet these criteria in some experiments (using the RIPPER classification algorithm), but generally, this is not true. PlanMiner-N on the other hand does, at least under certain noise levels. 

In terms of time efficiency, PlanMiner-N is 2\% slower than PlanMiner, even outperforming it in some experiments. This is because, in the face of noise, PlanMiner may spend much more time trying to fit an arithmetic expression (even using all the time allowed for this and reaching the timeout threshold), which compensates for the time spent by PlanMiner-N in applying the different anti-noise processes implemented. 

\section{Conclusions}
This paper presents a work about including a series of methods to the domain learning algorithm PlanMiner in order to increase its resistance to noise in the input data. The resultant algorithm is called PlanMiner-N and the new methods that implements are based on the preprocessing of the input data and the post-processing of the classification models used during the learning process of PlanMiner. The preprocessing of the input data is based on the discretization of noisy numerical data to automatically detect and handle outliers and to group similar values, while the postprocessing aims to unify the information contained in the learning models.

Results showed that these new methods increase notably the resistance of PlanMiner-N to noise. PlanMiner-N outperforms the original algorithm as well as the reference algorithms used in the experiments. These experiments were carried out using two different sets of input datasets and modifying them with three types of noise. 

The authors are exploring other methods to improve the domain learning algorithm. As PlanMiner (and now PlanMiner-N) disassociates the actions of the plans used as input when creating the datasets which later are used in the learning process, the authors are considering the option of design a new procedure able to take the contextual information contained in the order of the actions in the input plans and include it in the learning process in order to improve its performance. Another improvement to the PlanMiner algorithms that is being considered is the learning of more complex planning domains such as planning domains with conditional or stochastic actions.

\section{Acknowledgements}
This research is being developed and partially funded by the Spanish MINECO R\&D Project RTI2018-098460-B-I00

\bibliographystyle{elsarticle-num} 
\bibliography{bibliografia}

\end{document}